\documentclass{article}

\PassOptionsToPackage{numbers,compress}{natbib}
\usepackage[preprint]{neurips_2026}

\usepackage{graphicx} % Required for inserting images
\graphicspath{{figs/}}
\usepackage{subcaption}
\usepackage{amsmath, amssymb}
\usepackage{bm}          % bold math symbols, if used elsewhere
\usepackage{enumitem}    % for  \begin{enumerate}[label=()]
\usepackage[disable]{todonotes}
\usepackage{wrapfig}

\usepackage[hyphens]{url}            % simple URL typesetting
\usepackage[colorlinks=true,citecolor=orange,linkcolor=blue,breaklinks=true]{hyperref}       % hyperlinks
\usepackage{multirow}
\usepackage{fvextra}
\usepackage{amsthm}
\usepackage[capitalize]{cleveref}
\usepackage{mathtools}
\usepackage{algorithm}
\usepackage{booktabs}
\usepackage{colortbl}
\renewcommand{\vec}{\bm}
\renewcommand{\Re}{\mathbb{R}}
\newcommand{\E}{\mathbb{E}}

\newcommand{\Ytarget}{\mathcal{Y}}
\newcommand{\Xtarget}{\mathcal{X}}
\newcommand{\Ycontext}{\mathcal{Y}^\text{data}}
\newcommand{\Xcontext}{\mathcal{X}^\text{data}}
\newcommand{\ytarget}[1][]{\vec{y}^{#1}}
\newcommand{\ytargethat}[1][]{\widehat{\vec{y}}^{#1}}
\newcommand{\xtarget}[1][]{\vec{x}^{#1}}
\newcommand{\ycontext}[1][]{\vec{y}^{\text{data}, #1}}
\newcommand{\xcontext}[1][]{\vec{x}^{\text{data}, #1}}

\title{LLM Flow Processes for Text-Conditioned Regression}

\author{
  Felix Biggs\thanks{Equal contribution. Correspondence to \texttt{felix.biggs@wayve.ai} and \texttt{samuel.willis@secondmind.ai}.} \\
  Secondmind \\
  Wayve \\
  \And
  Samuel Willis\footnotemark[1] \\
  Secondmind \\
}

\begin{document}

\maketitle

\begin{abstract}
Recent work has demonstrated surprisingly good performance of pre-trained LLMs on regression tasks (for example, time-series prediction), with the ability to incorporate expert prior knowledge and the information contained in textual metadata.
However we observe major error cascades even in short sequences \(\lesssim 100\) points; these models are also computationally intensive and difficult to parallelise.
Marginal LLM predictions do not suffer this issue and are trivially parallelised, but can predict over-broad densities.
To address this, we propose combining these densities with a lightweight (diffusion-based) neural process.
We show that this combination leads to better-calibrated predictions overall, outputs locally consistent trajectories, and leads to text-conditioned function space selection in the meta-learner.
As part of this work we propose a gradient-free (and non-Monte Carlo) method for sampling from a product-of-experts of a score model and an `expert' (here the LLM predictive densities).
We believe this general method is of independent interest as it is applicable whenever an expert can be convolved with a Gaussian in closed form.
\end{abstract}

\section{Introduction}

Incorporating prior knowledge into regression models remains a fundamental challenge, particularly in domains where expert intuition is difficult to formalize.
Pre-trained large language models (LLMs) integrate an enormous amount of existing information and can perform many new real-world tasks from very few examples \citep{gpt3}.
Exciting new approaches are applying the implicit prior expert knowledge of these models to classical statistical tasks like regression, with some success \citep{LMPriors,Gruver2023-ul,llm-secretly-regressor,context-is-key,LLMP}.
Among these, the LLM-Process (LLMP; \citealp{LLMP}) can not only learn few-shot regression models, but can improve predictions with additional textual conditioning.
As an example, they demonstrate that predictions on a weather time series improve by adding the textual conditioning of ``daily temperature'', and further with the addition of ``in Montreal''.

\begin{figure*}[t]
    \centering
    \includegraphics[width=\textwidth]{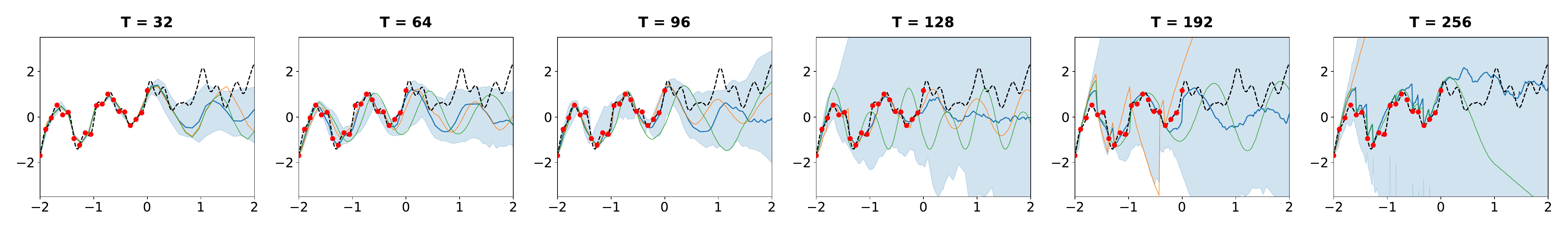}
    \includegraphics[width=0.6\textwidth]{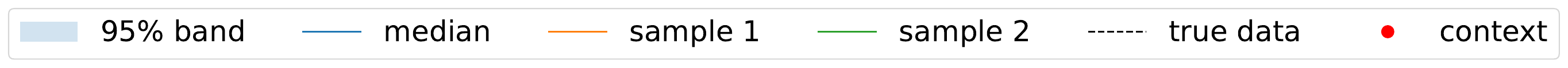}
    \caption{Predictions of the autoregressive Direct Prompt (\emph{LLM-Direct}) method on a test function, conditioned on context data (red) and prompt ``A periodic function with a linear trend'', at different numbers of predictive points \(T \in \{32, \dots, 256\}\); prediction trajectories are unrolled \(80\) times.
    By \(T=96\) some divergent trajectories begin to appear and by \(T=256\) these are the default causing even a divergent median prediction.
    It is visible that these divergent error cascading trajectories often commit to a certain linear trend as a kind of attractor state, even when this disagrees with the data; at \(T=256\) some trajectories ignore all but the first few data points.
    We note similarity to the phenomenon of neural text degeneration \citep{nucleus-sampling-20,DBLP:conf/nips/XuLY0LL22,DBLP:conf/iclr/WelleckKRDCW20,DBLP:conf/acl/WangLY0025}.
    }
    \label{fig:sweep}
\end{figure*}

However, this approach has drawbacks.
\citet{context-is-key} evaluate the LLMP and their own improved Direct Prompt method, and observe what they term `significant failures' on certain tasks, with predictions off by more than 500\%.
We observe a further failure mode of autoregressive models in this scenario (see, for example, \Cref{fig:sweep}) where increasing the number of points the LLM is predicting at (the horizon or precision in \(x\)) causes cascading failures.
We believe this relates to exposure bias when conditioning on their own predictions, which can cause exponential decay in predictive accuracy \citep{bengio-scheduled-sampling-15,lecun2023autoregressive,nucleus-sampling-20}.
Further, these models are computationally very expensive as the models are large and require sequential sampling of one point at a time, and the samples themselves depend on the order in which new points are predicted.

A complementary line of work tackles the prior-knowledge problem through meta-learning, for example the Neural Process \citep{neural-process,conditional-neural-p} approach, which learns a probabilistic distribution over regression functions from a set of `representative' datasets.
These datasets can be synthetically-generated using mechanisms (for example, causal graphs and augmentations) which mirror the processes generating real data.
Impressive real-dataset results have been achieved with transfer from such synthetic data, for example by \citet{transformers-bayesian-inference}.
Diffusion generative modelling-based \citep{Sohl-DicksteinW15,diffusion-ho,Song} approaches to this setting like Neural Diffusion Processes (NDPs; \citealp{NDP}) can learn intricate local structures while avoiding autoregression and being easy to parallelise \citep{flow-np}.
Despite the power of this approach which enables us to learn arbitrary meta-learners through data, encoding expert knowledge about the specific domain (function space selection) of a \emph{new} `test' dataset remains difficult, and auxiliary meta-data which could inform models (for example natural language descriptions of variables) are instead ignored.

The weaknesses of the above approaches are complementary: autoregressive LLMs suffer sampling pathologies leading to a lack of robustness, while NDPs have no way to incorporate textual information.
In order to combine the best strengths of each approach, we propose a new product-of-experts \citep{hinton-products-experts} sampling method to combine LLM Processes (with the ability to condition on prior world knowledge and rich, unstructured text) with NDPs (which can meta-learn over many divergent prior sources of regression data).

Our main contributions are as follows:
\begin{itemize}[nosep]
    %\item We move NDPs to a flow matching paradigm, more sophisticated conditioning and improve training with a scalable architecture.
    \item We introduce a principled gradient-free (with respect to the expert) non-Monte-Carlo product-of-experts sampling method for diffusion and flow generative models; this method is generally applicable whenever the expert(s) convolve in closed-form with a Gaussian.
    \item The product of LLM Flow Processes are an instantiation of this idea: by intersecting the NDP's meta-learned functions with the LLM's text-conditioned probabilities, we subselect the NDP-plausible trajectories consistent with the text.
    \item We empirically demonstrate failures of autoregressive LLM regression under increasing prediction resolution or horizon, and that LLM-FPs generate samples from their NDP generative distribution consistent with their textual conditioning.
\end{itemize}

\textbf{Structure.}
In \cref{sec:background} we discuss prerequisite and related work in neural processes, NDPs, diffusion, and diffusion model conditioning;
\cref{sec:method} introduces our general gradient-free method for expert-guided diffusion;
\cref{sec:method-llmfp} discusses its instantiation as LLM-FP, sampling regression functions with LLM expert advice;
\cref{sec:evaluation} describes our empirical evaluation methodology and results for these models, which are discussed in \cref{sec:discussion};
\cref{sec:conclusion} concludes with implications and future work.

\section{Background and Related Work}\label{sec:background}

\subsection{Meta-Learning of Probabilistic Regression}

The goal of meta-learning and Neural Processes \citep{neural-process}, which subsumes methods like NDPs \citep{NDP} and LLMPs \citep{LLMP}, is to obtain predictive distributions over new points conditioned on a set of context points (or `dataset', hence the `meta' modifier).
Formally, we have a sequence of \(n\) context input points \(\Xcontext = (\xcontext[i])_{i=1}^n\) and corresponding output points \(\Ycontext = (\ycontext[i])_{i=1}^n\) forming our dataset \(\mathcal{D} = (\Xcontext, \Ycontext)\).
We would like to predict \(m\) target outputs \(\Ytarget = (\ytarget[j])_{j=1}^m\) at their inputs \(\Xtarget = (\xtarget[j])_{j=1}^m\).
Specifically, we would like to elicit or learn the predictive distribution \(p(\Ytarget| \Xtarget, \mathcal{D}).\)
Note that we may wish to learn with a parametrised distribution which works for any possible \(m\) and \(n\),\footnote{This problem has close relationships to learning a distribution over functions or stochastic process, if we satisfy the conditions of the Kolmogorov Extension Theorem it can define a valid stochastic process (see \cref{appendix:KET}).} and have the option to condition also on additional side-information, \(\mathcal{C}\), like meta-data.

\textbf{LLM Processes and Direct Prompt.}\label{sec:llmp}
LLM Processes (\citealp{LLMP}; which build on previous work such as \citealp{LMPriors,Gruver2023-ul,llm-secretly-regressor}) elicit a predictive distribution like that above by repurposing LLM predictive probabilities.
Additional contextual information can simply be incorporated into the prompt, utilising the `expert knowledge' of the LLM.
The context data \(\mathcal{D}\) and target locations \(\Xtarget\) are contained in the prompt.

Samples are drawn by unrolling the LLM decoder in an autoregressive way (termed A-LLMP in \citealp{LLMP}).
Note that the predictive probabilities will depend on the order we choose, which makes full density estimation infeasible.
This scenario is examined by \citet{context-is-key} who introduce an improved Direct Prompt method with constrained decoding that significantly improves on LLMP results.
However, they note in all such models a prevalence of what they term `significant failures', where predictions are off by more than 500\%.
We additionally run into standard issues with autoregressive models (where the model is conditioning on its own imperfect outputs which do not resemble training data \citealp{bengio-scheduled-sampling-15}): errors can compound, possibly exponentially with longer prediction horizons (see \cref{fig:sweep}); predictions can worsen dramatically with the number of points we wish to predict.
We note in particular a tendency of mode collapse to a kind of `attractor state' of predicting linear trends, perhaps related to the neural text degeneration phenomenon or `repetition curse' \citep{nucleus-sampling-20,DBLP:conf/nips/XuLY0LL22,DBLP:conf/iclr/WelleckKRDCW20,DBLP:conf/acl/WangLY0025}.
\citet{lecun2023autoregressive} argues that these problems fundamentally limit decoder-based LLM models.

An alternative to these autoregressive models is the \emph{independent marginal} formulation (I-LLMP), where we assume conditional independence of the target points.
The probabilities of each \(\ytarget[j]\) can then directly obtained through the output logits of the LLM, with joint probability via multiplication \citep{LLMP}.
We note however this may not capture correlation in the evaluation points, and we observe that lack of reasoning about the local structure can lead to some strange outliers (see the LLM-Independent \emph{median} in \cref{fig:changepoint}).

\textbf{Neural Diffusion Processes.}\label{sec:ndp}
\citet{NDP} instead focuses on learning a neural (diffusion) process defined by a \emph{diffusion} generative model \citep{Sohl-DicksteinW15,diffusion-ho,Song} in the output variables \(\Ytarget\).
Samples can then be drawn from the joint distribution via a diffusion forward process.
By using an architecture that is equivariant to permutation of inputs and outputs (for example, by using self-attention) this produces exchangeability (in the sense of the KET; \cref{appendix:KET}), enabling coherent draws.
The original paper used a de-noising score matching loss \citep{denoising-score-match} taking input variables (\(\Xcontext, \Xtarget\)) to learn a distribution over both target \emph{and} context outputs (\(\Ycontext, \Ytarget\)); conditioning on context outputs was done via a Re-Paint style recycling of context points \citep{repaint}.
Here we update conditioning to a modern method (based on \citealp{pseudoinverse-song,inverse-flows-pokle}, see \cref{sec:conditional}; this also allows us to account for standard Gaussian measurement noise), a flow matching objective \citep{flow-matching-albergo,flow-matching-lipman,flow-straight-fast}, and a more standard diffusion transformer \citep{diffusion-transformer-23}.
We note similar works of \citet{flow-np} and \citet{DataAssim}, which respectively define an flow-matching-based NDP with \emph{context} output points as inputs rather than diffusion variables, and generate trajectories with a diffusion model.

\textbf{Synthetic Data.}
A significant component of the Neural process literature looks at generating synthetic data for training models before transfer to real-world datasets \citep{DBLP:journals/corr/abs-2101-03606,DBLP:journals/corr/abs-2108-09676,DBLP:conf/icml/0005RHRH25,DBLP:journals/corr/abs-2510-21204,DBLP:journals/corr/abs-2602-11139,DBLP:journals/corr/abs-2512-03307}.
For example \citet{NDP} trained on Gaussian Process samples, and \citet{TabPFN2023} trained on data generated using random causal graphs and special augmentations (with discrete labels).
\citet{transformers-bayesian-inference} notes that this is equivalent to learning to implicitly marginalise over a Bayesian posterior distribution with the prior defined by the data-generation process, as also examined by \citet{NDP} (see in particular their Figure 3).
Improved data-generation procedures have been explored in subsequent work \citep{tabpfn-nature-v2,DBLP:journals/corr/abs-2502-17361}, including adversarially-designed generators \citep{DBLP:journals/corr/abs-2502-04573}.
These approaches are complementary to our contribution and could be incorporated into our training pipeline to further strengthen the learned prior, though we do not pursue them here.

\subsection{Diffusion and Flow Matching}

Diffusion models \citep{Sohl-DicksteinW15,diffusion-ho,Song} and their generalisation to flow matching (\citealp{flow-matching-albergo,flow-matching-lipman,flow-straight-fast}; see \citealp{flow-matching-guide} for a comprehensive introduction) provide a simulation-free learning objective for generative modelling via differential equations.
We adopt the popular class of affine conditional flows interpolating from Gaussian noise \(\vec{Y}_0 \sim p_0 = p_\text{noise}\) to data \(p_\text{data} = p_1\),\footnote{We notate the diffused variable as \(\vec{y}\) or \(\vec{Y}\), rather than the more usual \(\vec{X}\), as it will also represent the dependent variable in regression as defined by an NDP.} which can be defined by the conditional probability paths (for some scheduler pair \(\alpha_t, \sigma_t\)\footnote{In practice, we use the conditional Optimal Transport (OT) path \citep{flow-matching-lipman}: \(\alpha_t = t\) and \(\sigma_t = 1 - t\).}):
\[ p_{t|1}(\vec{y}|\vec{y}_1) = \mathcal{N}(\vec{y}|\alpha_t \vec{y}_1, \sigma_t^2 I).\]
The flow matching or de-noising score matching objective \citep{denoising-score-match} can be used to learn a neural network approximation to the marginals \(p_t(\vec{y}) = \int p_{t|1}(\vec{y}|\vec{y}_1) p_1(\vec{y}_1) \mathrm{d}\vec{y}_1\).
This network can generally be parametrised as a probability flow for sampling, or equivalently (in the affine Gaussian case) a `score network' \(s_t^\theta(\vec{y}_t) \approx \nabla_{\vec{y}_t} \log p_t(\vec{y}_t)\).
These can be linearly converted between for sampling purposes, as can direct prediction (via Tweedie's formula) of the expected final output \(\widehat{\vec{y}}_{1|t}(\vec{y}_t) \coloneq \E[\vec{Y}_1|\vec{Y}_t = \vec{y}_t]\); see \cref{appendix:flow-matching} for more details.

\textbf{Conditioning Diffusion Models.}\label{sec:conditional}
Using our approximated flow or score function \(s^\theta_t\), we can adapt the above to sampling from a conditional distribution \(p_1(\vec{y}|\vec{z})\) for some conditioning variable \(\vec{z}\) (for example, we could pre-specify certain components of \(\vec{y}\) for an infilling task).
From Bayes' rule,
\[ \nabla_{\vec{y}_t} \log p_t(\vec{y}_t|\vec{z}) = \nabla_{\vec{y}_t} \log p_t(\vec{y}_t) + \nabla_{\vec{y}_t} \log p_t(\vec{z} | \vec{y}_t).\]

\citet{inverse-flows-pokle} and \citet{pseudoinverse-song} consider problems where the measurement model \(p_\text{meas}(\vec{z}|\vec{y}_1)\) is linear in \(\vec{y}_1\) plus Gaussian noise.
Generalising a method of \citet{diffusion-inverse-chung}, they express
\[ p_t(\vec{z} | \vec{y}_t) = \int p_\text{meas}(\vec{z}|\vec{y}_1) \; p_{1|t}(\vec{y}_1|\vec{y}_t) \mathrm{d}\vec{y}_1, \]
and substitute the approximation;
\begin{equation}\label{eq:basic-approximation-p1}
p_{1|t}(\vec{y}_1 \mid \vec{y}_t) \;\approx\; \mathcal{N}\left(\vec{y}_1; \,\widehat{\vec{y}}_{1|t}(\vec{y}_t),\, r_t^{2}I\right).
\end{equation}
where \(\widehat{\vec{y}}_{1|t}(\vec{y}) \coloneq \E[\vec{Y}_1 | \vec{Y}_t = \vec{y}]\) is linearly derived from and approximated using our score network (see Eq.~\ref{eq:x1-from-flow} in \cref{appendix:flow-matching}), and \(r_t\) is a time-dependent noise value that can depend on the data, which \citet{inverse-flows-pokle} argue should be set to \(r_t^2 = \sigma_t^2 / (\sigma_t^2 + \alpha_t^2)\).
Under this approximation the score correction can be written in closed form and used to guide diffusion towards the desired conditional.
In practice because the measurement model is learned separately from the diffusion model, the calibration of the models may not match; an additional multiplicative weighting may be added to the conditioning score to compensate for this.
Using the correspondence between the score and the flow (Eq.~\ref{eq:flow-from-score}) this guidance for the score function can then be adapted into guidance via flow directly.
We note further proposed methods for improving the performance of conditional sampling, such as using variance estimates in the approximation \cref{eq:basic-approximation-p1} \citep{rozet-em-diffusion}, or aligning the prior density \(p_0\) using Gaussian Processes \citep{DBLP:conf/iclr/KolloviehLLSG25}.

Closely related work looks at sampling from energy tilted densities like \(\propto p_1(\vec{y}) e^{-\ell(\vec{y})},\) for energy (loss) \(\ell\) \citep{DBLP:conf/icml/SongZYM0KCV23,DBLP:conf/iccv/YuWZGZ23,DBLP:conf/nips/GuoYYCW24,DBLP:conf/nips/ShenJYWHL24,DBLP:conf/iclr/BansalCSSGGG24,Kong2024-gx}.
These works also use a guided score function for this combined density and make the same normal approximation for \(p_{1|t}\), leading to an additional score term like \(\nabla_{\vec{y}_t} \log \int \mathcal{N}\left(\vec{y}_{1|t}; \,\widehat{\vec{y}}_1(\vec{y}_t),\, r_t^{2}I\right) e^{-\ell(\vec{y}_1)} \mathrm{d}\vec{y}_1,\) which is then approximated in some way, for example through Monte Carlo sampling.
This can be used for a variety of purposes, for example by \citet{Kong2024-gx} to optimise an objective \(\ell\) on complex domains learned through the diffusion model while avoiding infeasible solutions like non-synthesizable molecules.
A similar additional score term appears in our gradient-free product-of-experts method introduced in \cref{sec:method}.

\section{Gradient-Free Product of Experts with Diffusion}\label{sec:method}

We utilise a products of experts \citep{hinton-products-experts} which is the natural way to combine two distributions encoding independent sources of information.
We begin with a distribution \(p_1\) defined by our diffusion model,\footnote{For simplicity in this section we consider the standard setting of diffusion in single variable \(\vec{y}_t\), rather than the more complex NDP setup.} and \(q\) an ``expert'' distribution (for example LLMP predictions) we wish to integrate information from via an extra conditioning variable \(c\) (for example a textual prompt).
We define the product of experts distribution of these as
\[ \pi_1(\vec{y}_1|c) \coloneq \frac{1}{Z_c} q(\vec{y}_1|c) p_1(\vec{y}_1)\]
where \( Z_c = \int q(\vec{y}_1|c) p_1(\vec{y}_1) \mathrm{d}\vec{y}_1\) is a normalising constant.
An advantage of this product-of-experts formulation is that density is concentrated where both \(q\) and \(p_1\) have probability mass, so we are unlikely to end up in the low-density regions of \(p_1\) where our score model has poorly learned the underlying density.

Using the same conditional probability distributions as our original flow model (for example, the OT affine Gaussian flow), we define a time-dependent version of the product distribution,
\begin{align*}
\pi_t(\vec{y}_t|c) \coloneq \int \pi_1(\vec{y}_1) \,p_{t|1}(\vec{y}_t \mid \vec{y}_1)\,\mathrm{d}\vec{y}_1
= \frac{1}{Z_c} \int q(\vec{y}_1|c) p_1(\vec{y}_1)\,p_{t|1}(\vec{y}_t \mid \vec{y}_1)\,\mathrm{d}\vec{y}_1.
\end{align*}

In order to produce guided sampling we use Bayes' Theorem to rewrite the score function
\begin{align*}
\nabla_{\vec{y}_t}\,\log\pi_t(\vec{y}_t)
  &= \nabla_{\vec{y}_t}\log\!\int q(\vec{y}_1|c)\,p_{t|1}(\vec{y}_t \mid \vec{y}_1)\,p_1(\vec{y}_1)\,\mathrm{d}\vec{y}_1 \\
  &= \nabla_{\vec{y}_t}\log\!\int q(\vec{y}_1|c)\,p_{1|t}(\vec{y}_1 \mid \vec{y}_t)\,p_t(\vec{y}_t)\,\mathrm{d}\vec{y}_1 \\
  &= \nabla_{\vec{y}_t}\log p_t(\vec{y}_t)
     + \nabla_{\vec{y}_t}\log\!\int q(\vec{y}_1|c)\,p_{1|t}(\vec{y}_1 \mid \vec{y}_t)\,\mathrm{d}\vec{y}_1.
\end{align*}
This closely parallels the conditional score derivation in \cref{sec:conditional}, with the expert \(q(\cdot|c)\) playing the role of the measurement model.
We also adopt the approximation of \(p_{1|t}(\vec{y}_1 \mid \vec{y}_t) \;\approx\; \mathcal{N}\left(\vec{y}_1;\, \widehat{\vec{y}}_{1|t}(\vec{y}_t),\, r_t^{2}I\right)\) as in \cref{eq:basic-approximation-p1}, to find
\begin{equation}\label{eq:conv-density}
  \int q(\vec{y}_1|c)\,p_{1|t}(\vec{y}_1 \mid \vec{y}_t)\,\mathrm{d}\vec{y}_1 \approx \int q(\vec{y}_1|c)\, \mathcal{N}\left(\vec{y}_1;\, \widehat{\vec{y}}_{1|t}(\vec{y}_t), r_t^{2}I\right) \, \mathrm{d}\vec{y}_1 \eqcolon \widetilde{q}_{r_t}(\widehat{\vec{y}}_{1|t}(\vec{y}_t)|c)
\end{equation}
also using \(\widehat{\vec{y}}_{1|t}(\vec{y}) \coloneq \mathbb{E}[\vec{Y}_1| \vec{Y}_t = \vec{y}]\) as defined in \cref{sec:conditional}.
Similarly, \(r_t\) is chosen in some way based on the data, for example following \citet{inverse-flows-pokle}.
Sampling is then possible through the score formulation of diffusion via
\begin{equation}\label{eq:moe-score}
\nabla_{\vec{y}_t} \log \pi_t(\vec{y}_t|c)
\approx s_t^\theta(\vec{y}_t) + \nabla_{\vec{y}_t} \log \widetilde{q}_{r_t}(\widehat{\vec{y}}_{1|t}|c).
\end{equation}

\(\tilde{q}\) can be rewritten as a convolution with a zero-mean Gaussian, and hence corresponds to a density, which is evaluated at \(\widehat{\vec{y}}_{1|t}\).\footnote{Since \(
\widetilde{q}_{r_t}(\vec{y}) = \int q(\vec{y}') \,\mathcal{N}(\vec{y}';\, \vec{y}, r_t^2 I) \, \mathrm{d}\vec{y}'
= \int q(\vec{y}') \,\mathcal{N}(\vec{y}' - \vec{y}; 0, r_t^2 I) \, \mathrm{d}\vec{y}'
\).}
Specifically, it is the density of draws from \(q\) corrupted by Gaussian noise of scale \(r_t\) - a quantity arising commonly in signal processing (for example, as a ``Gaussian filter'') or in kernel density estimation.
\(\widetilde{q}_{r_t}\) is infinitely differentiable, and its score function can be computed without \emph{taking gradients of \(q\)}, only through the score network via \(\widehat{y}_{1|t}\); in this sense our method is \emph{gradient free}.

Up to \cref{eq:moe-score} our method is similar to the conditional score \citep{inverse-flows-pokle,pseudoinverse-song} or to energy-tilted methods \citep{DBLP:conf/icml/SongZYM0KCV23,DBLP:conf/iccv/YuWZGZ23,DBLP:conf/nips/GuoYYCW24,DBLP:conf/nips/ShenJYWHL24,DBLP:conf/iclr/BansalCSSGGG24} which use Monte Carlo estimation for the second term.
However unlike these works we will instead work with distributions where the convolution is actually tractable, noting that a large such class exists.
For example, in the case of an LLM (as considered below) convolution is trivial because it outputs probabilities in bins (see Eq.~\ref{eq:q-bins}).
This convolution trick avoiding gradients of \(q\) is a core contribution, and could be extended in other settings: for example, all piecewise polynomial functions have closed form convolutions with Gaussians using only polynomials and \(\Phi\); this takes us into the field of spline smoothing \citep{splines99}.

\section{LLM-Flow Processes: Diffusing Regressors with LLM Expert Guidance}\label{sec:method-llmfp}

We use an NDP as defined in \cref{sec:ndp} to model a joint distribution, and add product-of-expert conditioning \(q\) via the method of the I-LLMP. In this case
\(q\) factorises into a set of conditional marginals, which allows each LLM expert to focus on only predicting one point.
This factorisation of \(q\) implies factorisation of \(\widetilde{q}\) (due to the diagonal covariance in Eq.~\ref{eq:conv-density}):
\begin{align*}
q(\Ytarget|\Xtarget, \mathcal{D}) = \prod_{i=j}^m q(\ytarget[j]|\xtarget[j], \mathcal{D})
\implies
\quad \widetilde{q}_{r_t}(\Ytarget|\Xtarget, \mathcal{D}) = \prod_{j=1}^m \widetilde{q}_{r_t}(\ytargethat[j]_{1|t} | \xtarget[j], \mathcal{D}).
\end{align*}
We here defined \(\ytargethat[j]_{1|t}(\mathcal{Y}') \coloneq \mathbb{E}[\ytarget[j]_1 | \Ytarget_t = \mathcal{Y}', \Xtarget, \mathcal{D}]\), \emph{i.e.} the expected value of a particular output via the score network Tweedie estimate, which depends simultaneously on all the \(\vec{y}_t\) values at the current time (because all target outputs are simultaneously diffused, capturing joint dependencies).

From this we derive a score function like that in \cref{eq:moe-score} for sampling of product-of-expert distribution \(\pi_1\) via
\begin{equation*}
\nabla_{\Ytarget_t} \log \pi_t(\Ytarget_t|\Xtarget, \mathcal{D})
\approx s_t^\theta(\Ytarget_t; \Xtarget, \mathcal{D})
+ \sum_{j=1}^m \nabla_{\Ytarget_t} \log \widetilde{q}_{r_t}(\ytargethat[j]_{1|t}|\xtarget[j], \mathcal{D}).
\end{equation*}

To calculate the factorised convolved distributions \(\widetilde{q}_{r_t}\), we note that I-LLM outputs probability in bins, so the integral can be reformulated as a sum over disjoint bins \(B \in \mathcal{B}, B \subset \Re^{\text{out}}\) spanning the space.
In the simplest one-dimensional case, \(q\) is piecewise constant on a series of intervals \(B = [a_B, b_B]\) for \(B \in \mathcal{B}\) partitioning \(\Re\) with density values \(q_B\).
The smoothed density is then
\begin{equation}\label{eq:q-bins}
\widetilde{q}_{r}(y) = \sum_{B \in \mathcal{B}} q_B \left[ \Phi\left(\frac{b_B - y}{r}\right) - \Phi\left(\frac{a_B - y}{r}\right)\right],
\end{equation}
using \(\Phi\), the standard Gaussian cumulative distribution function.

Note that the score network \(s_t^\phi\) here can itself be an unconditional NDP with score-based conditioning, or a directly conditional model such as that of \citet{flow-np}.
With their equivariant architectures our implied product-of-experts probability distribution is also equivariant under re-ordering of \((\xtarget[j], \ytarget[j])\) target pairs.
This equivariance in context predictions is not shared by the autoregressive A-LLMP, since predictions will depend on the rollout ordering.

\section{Empirical Evaluation}\label{sec:evaluation}

\begin{figure*}[t]
    \centering
    \includegraphics[width=\textwidth]{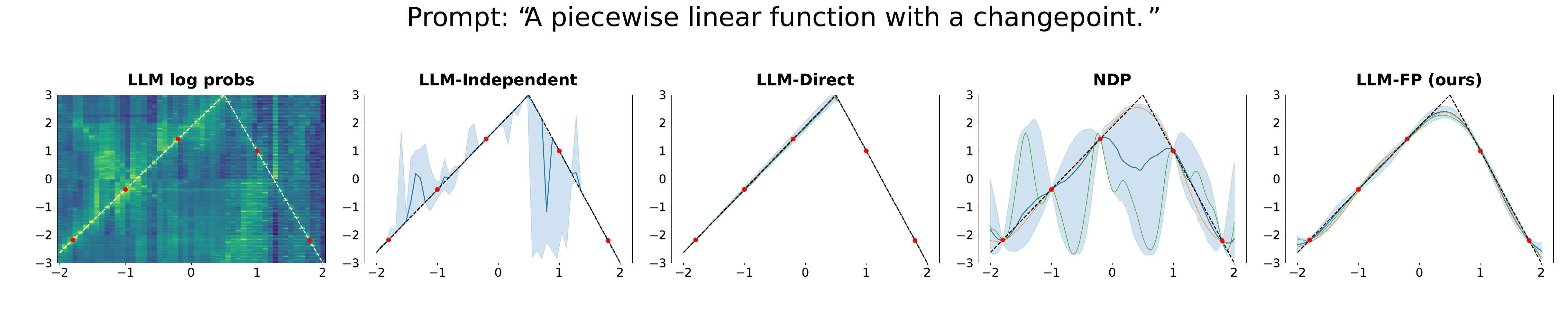}
    \includegraphics[width=0.6\textwidth]{figs/figure_legend_with_truth.pdf}
    \caption{Change-point regression with N=6 context points. Methods with access to the textual change-point description (LLM-Direct, LLM-Independent, LLM-FP) recover the discontinuity; the unconditional NDP reverts to its periodic/RBF prior. Two trajectory samples are shown for methods that produce coherent samples (LLM-Direct, NDP, LLM-FP); the LLM marginal density used by LLM-Independent and LLM-FP is shown on the left.}
    \label{fig:changepoint}
\end{figure*}

\begin{figure*}[t]
    \centering
    \includegraphics[width=\textwidth]{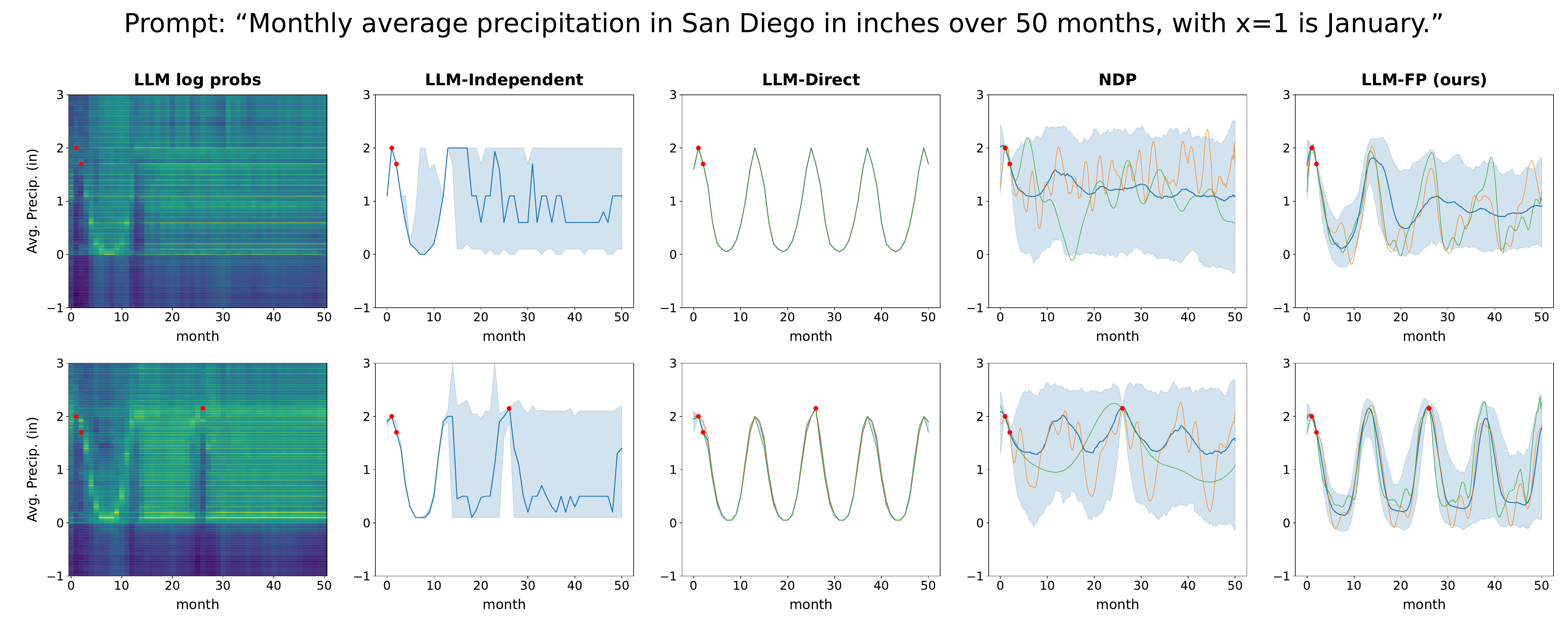}
    \includegraphics[width=0.47\textwidth]{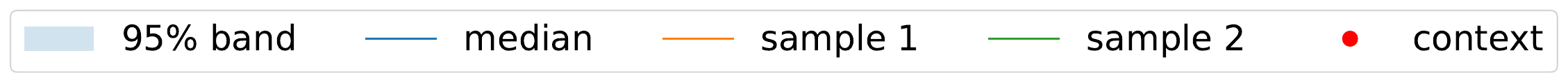}
    \caption{Synthetic San Diego rainfall extrapolation with N=2 (top) and N=3 (bottom) observed values and a shared textual prompt describing seasonal structure. LLM-Direct collapses to a near-deterministic seasonal template in both cases, largely insensitive to the additional observation. The independent LLMP assigns broad marginal mass beyond the first year. The LLM-FP concentrates on periodic trajectories of approximately the correct amplitude and period; adding a third observation further narrows the inferred period and amplitude.}
    \label{fig:sandiego}
\end{figure*}

\begin{figure*}[t]
    \centering
    \includegraphics[width=\textwidth]{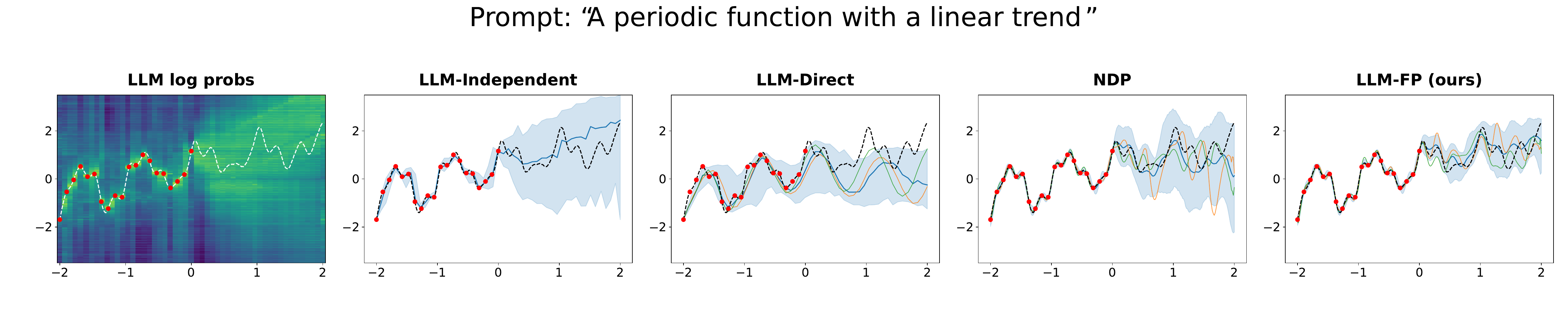}
    \includegraphics[width=0.6\textwidth]{figs/figure_legend_with_truth.pdf}
    \caption{Extrapolation on a periodic function with an added linear trend, with N=20 context points in the range [-2, 0]. The periodic component is consistent with the NDP training prior but the linear trend is not: the NDP remains well-calibrated but reverts towards its prior mean away from the observations. LLM-Direct fails to fit the context and the true function lies outside its predictive distribution. The independent LLMP captures the upward trend but with very broad marginals. LLM-FP follows the trend while retaining narrower uncertainty than the independent LLMP.}
    \label{fig:periodic_with_linear_trend}
\end{figure*}  % todo: remove to appendix if no space

\begin{figure*}[t]
    \centering
    \includegraphics[width=\textwidth]{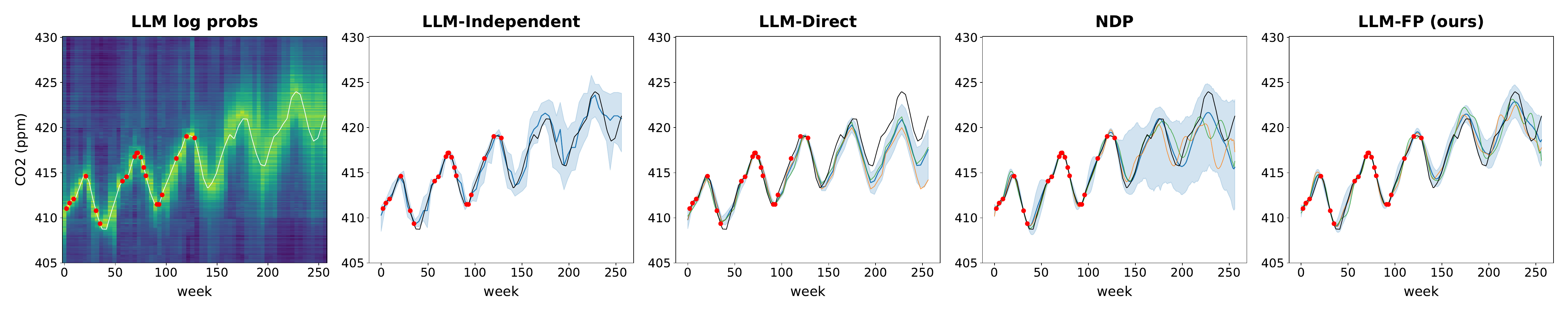}
    \includegraphics[width=0.6\textwidth]{figs/figure_legend_with_truth.pdf}
    \caption{Mauna Loa atmospheric CO\textsubscript{2} measurements, with N=20 randomly selected weeks from 2019-01-06 to 2021-06-13 as context (full prompt in Appendix~\ref{appendix:prompt}). The independent LLMP marginals are approximately correct with broad pointwise uncertainty; NDP trajectories are locally smooth but drift from the true function in phase and amplitude over the window; LLM-Direct is sharply overconfident (NLPD is severely inflated, see Table~\ref{tab:results}). LLM-FP combines the two, staying close to the true function throughout.}
    \label{fig:mauna_loa_weekly}
\end{figure*}

%\todo[inline]{bug fig 1? - 50 percentile line isn't strictly inside the other lines?}

%\todo[inline]{better to replace NDP+LLM conditioning with just LLM but also at ycontext - q(ycontext|D), clearly it has low variance there, more straightforward.}

\textbf{NDP base model.}
We use a modernised NDP with flow matching objective, diffusion transformer backbone and Gaussian conditioning \citep{inverse-flows-pokle} in place of the SDE-based diffusion and in-paint conditioning of \citet{NDP}; see \cref{appendix:additional-training} for further details.
We found this modernisation, which gave substantial improvements, essential for coherent conditional and unconditional generation
(\emph{cf.} for example Fig.~3 in NDP showing samples that clearly do not match their training distribution versus \cref{fig:unconditional_ndp,fig:conditional_ndp} in \cref{appendix:validation}).
We note this improved \emph{NDP} provides a considerably more challenging baseline for comparison than the original model.
Like \citet{NDP} we train on a simple synthetic dataset of Gaussian Process samples of mixed kernel types and length-scales (somewhat richer than in the above, for example, including periodic kernels, see \cref{appendix:additional-training} for details); the NDP implicitly marginalizes over these and gives a Bayesian posterior prediction based on this data as a prior \citep{transformers-bayesian-inference}.

\textbf{Baselines.}
We compare LLM-FP against three baselines that isolate different design choices in this space: our improved NDP without LLM conditioning (\emph{NDP}); an autoregressive LLM predicting one point at a time (\emph{LLM-Direct}); and independent LLM predictions (\emph{LLM-Independent}).

For the autoregressive \emph{LLM-Direct}, we follow the Direct Prompt constrained generation method of \citet{context-is-key}.
We also compute independent LLM probabilities using binning, similarly to \citet{LLMP}.
The binned LLM log-probabilities used by LLM-FP are identical to the marginals sampled by \emph{LLM-Independent}; we display these in figures as \emph{LLM log probs}.
\emph{LLM-Independent} is sampled directly from these.

All LLM-based methods (LLM-Direct, LLM-Independent, LLM-FP) use the Qwen-3.6-27B LLM \citep{qwen36_27b}.
We add a reasoning block before predictions for all LLM-based methods which we detail in Appendix~\ref{appendix:prompt}.
These methods use identical prompts within each scenario; prompts are not tuned per method.
Note that we did not do any fine-tuning of our LLMs.

Across scenarios, the autoregressive \emph{LLM-Direct} predicts at 51–64 points as does \emph{LLM-Independent}; this is a favourable regime for \emph{LLM-Direct}, since autoregressive performance degrades severely beyond roughly 100 points (\cref{fig:sweep}).
NDP and LLM-FP predict at 256 points throughout without difficulty; LLM-FP uses LLM evaluations only at the same points as \emph{LLM-Independent}, since the NDP provides the remaining structure, representing a \({\sim}4\times\) computational saving over LLM predictions at all 256 points.

\textbf{Metrics.} In all evaluations with a ground truth we include the mean absolute error (MAE), the continuous ranked probability score (CRPS) which captures some degree of calibration, and the negative log likelihood of per-point fitted Gaussians (NLPD), which better captures overconfidence.

\textbf{Scenarios.} We evaluate across four scenarios: a simple change-point function to test text faithfulness (\cref{fig:changepoint}); rainfall extrapolation (\cref{fig:sandiego}) as used in \citep{LLMP}; a synthetic periodic with trend function (\cref{fig:periodic_with_linear_trend}); and a real-data problem from the Mauna Loa atmospheric dataset \citep{maunaloa2026trends} (\cref{fig:mauna_loa_weekly}).

\textbf{Results.}
Across the four scenarios (Figs. \ref{fig:changepoint}–\ref{fig:mauna_loa_weekly}), LLM-FP produces trajectories that are simultaneously coherent (unlike LLM-Independent) and faithful to the textual context (unlike NDP), avoiding both the autoregressive collapse of LLM-Direct and the broad marginals of LLM-Independent.
Quantitatively, LLM-FP achieves the best or near-best CRPS and NLPD on Periodic+linear and Mauna Loa~\citep{maunaloa2026trends} (Table~\ref{tab:results}); on the text-dominated Change point problem, LLM-FP smooths out the occasional outliers in the LLM-Independent samples through the NDP's local consistency, but does not match LLM-Direct's near-perfect reproduction of the discontinuity.

\begin{table}[t]
\centering
\caption{Predictive performance across regression tasks; lower is better, best per row in bold. Cells are central $\pm$ half-width of the 95\% bootstrap CI over evaluation positions. On the text-dominated Change point problem, the autoregressive LLM-Direct produces near-perfect single-trajectory predictions and LLM-FP does not match its accuracy; LLM-FP nonetheless improves substantially over the marginal LLM-Independent baseline on every metric, and is the best method on Periodic + linear and Mauna Loa.}
\label{tab:results}
\footnotesize
\setlength{\tabcolsep}{4pt}
\begin{tabular}{llcccc}
\toprule
Task & Metric & LLM-Independent & LLM-Direct & NDP & LLM-FP (ours) \\
\midrule
\multirow{3}{*}{Change point} & MAE & $0.29_{\pm.13}$ & $\mathbf{0.02_{\pm.00}}$ & $0.73_{\pm.10}$ & $0.09_{\pm.01}$ \\
 & CRPS & $0.12_{\pm.07}$ & $\mathbf{0.01_{\pm.00}}$ & $0.43_{\pm.05}$ & $0.08_{\pm.01}$ \\
 & NLPD & $0.06_{\pm.27}$ & $\mathbf{-1.84_{\pm.38}}$ & $0.77_{\pm.13}$ & $-0.40_{\pm.31}$ \\
\midrule
\multirow{3}{*}{Periodic + linear} & MAE & $0.44_{\pm.14}$ & $0.93_{\pm.26}$ & $0.41_{\pm.08}$ & $\mathbf{0.26_{\pm.03}}$ \\
 & CRPS & $0.37_{\pm.07}$ & $0.69_{\pm.21}$ & $0.29_{\pm.05}$ & $\mathbf{0.18_{\pm.02}}$ \\
 & NLPD & $1.00_{\pm.15}$ & $2.05_{\pm.73}$ & $0.70_{\pm.11}$ & $\mathbf{0.20_{\pm.09}}$ \\
\midrule
\multirow{3}{*}{Mauna Loa CO\textsubscript{2}} & MAE & $0.84_{\pm.24}$ & $1.91_{\pm.41}$ & $1.11_{\pm.18}$ & $\mathbf{0.80_{\pm.10}}$ \\
 & CRPS & $0.62_{\pm.16}$ & $1.52_{\pm.31}$ & $0.74_{\pm.11}$ & $\mathbf{0.56_{\pm.07}}$ \\
 & NLPD & $1.48_{\pm.18}$ & $4.18_{\pm.89}$ & $1.74_{\pm.09}$ & $\mathbf{1.43_{\pm.15}}$ \\
\bottomrule
\end{tabular}
\end{table}

\section{Discussion}\label{sec:discussion}

\textbf{Mechanism.} LLM-FP combines two complementary sources of information: the NDP is trained to provide a coherent stochastic process prior, producing smooth and time-consistent trajectories, while the LLM provides semantic information from the textual prompt. The product-of-experts conditioning concentrates probability mass on trajectories that are probable under both factors, with the result that LLM-FP samples are typically more concentrated than either component used in isolation (see, e.g., Figs.~\ref{fig:sandiego}, \ref{fig:periodic_with_linear_trend}). This is most useful in underdetermined settings, where neither the observed data nor the textual information alone identifies the relevant function class.
It additionally appears that the NDP component of the LLM-FP can provide some additional robustness by exploiting local structures; \cref{fig:changepoint} shows that even the median LLM-Independent prediction can have outlier predictions at certain input points, a problem the LLM-FP totally eliminates.

\textbf{Text as function-class selection.} A useful interpretation of the LLM conditioning is that it performs a form of soft function-class selection. The NDP prior marginalises over functions drawn from a mixture of periodic and RBF kernels; when the text describes seasonal or periodic behaviour, the LLM-derived probabilities suppress trajectories inconsistent with this description, effectively reweighting the NDP prior towards its periodic component. In Fig.~\ref{fig:periodic_with_linear_trend}, the LLM-derived probabilities favour upward-trending predictions that are not directly represented by the NDP training prior, allowing the combined model to extrapolate behaviour that lies near the boundary of the NDP support. In both cases, the text does not replace the process prior, but selects among, or extrapolates from, behaviours already assigned non-negligible probability by the NDP.

\textbf{Failure modes of direct LLM regression.} The autoregressive baseline (LLM-Direct) performs well on simple text-dominated problems especially when the function is well captured by its attractor state of linear trends (Fig.~\ref{fig:changepoint}) but is fragile when predictions must be unrolled over many output locations: in Fig.~\ref{fig:sandiego} it collapses to a near-deterministic seasonal template, and Fig.~\ref{fig:sweep} shows sample quality degrading progressively as the number of prediction points grows. The independent LLMP avoids these failure modes by treating each location separately, but consequently does not define coherent joint samples. LLM-FP uses the marginal probabilities as a conditioning signal while retaining the joint structure of the NDP, sidestepping both failure modes.

\textbf{Limitations.} The product-of-experts update implicitly treats the NDP and LLM factors as independent. If the two are strongly correlated, the update may double-count evidence and over-concentrate; this risk is exacerbated when the LLM-derived probabilities are miscalibrated or overconfident.
This does not appear to happen in practice, which we speculate is because the NDP and LLM are integrating different sources of knowledge for their predictions: the independent LLM integrates textual information at individual points, while the NDP leans on the local function structures contained in its trajectories.
More broadly, the quality of the combined model depends on both the calibration of the LLM marginals and on whether the NDP prior assigns sufficient mass to the relevant functions: if the desired behaviour lies far outside the NDP support, LLM conditioning can only reweight the existing prior rather than create new trajectory structure (this is one source of the LLM-FP shortfall on Change point in Table~\ref{tab:results}); conversely, if the LLM marginals are too diffuse, they provide only weak conditioning.
Our experiments use a single 27B-parameter open-weight LLM without fine-tuning and a simple synthetic NDP training distribution, and we do not study how performance scales with LLM size, prompt design, or richer NDP priors.

\section{Conclusion}\label{sec:conclusion}

%\todo[inline]{add that trajectories are actaully helpful for getting overall predictions - some kind of local structure from NDP is helpful - see experiments}

We introduced LLM-FP, a method for conditioning Neural Diffusion Processes on textual information through LLM-derived probabilities.
The approach preserves the coherent trajectory-level samples of NDPs while incorporating semantic information that is unavailable from numerical observations alone.
Across synthetic and real regression tasks, this enables text-conditioned predictions that remain smooth and time-consistent, avoiding the instability observed in direct autoregressive LLM regression. The NDP's local trajectory structure also regularises the LLM marginals themselves: pointwise predictions at neighbouring inputs may be individually well-calibrated yet mutually inconsistent, and coupling them through the NDP rules out such inconsistencies.

More broadly, LLM-FPs provide a way to combine synthetic-data-trained stochastic process priors with expert knowledge expressed in language.
This suggests a path towards regression models that retain the calibration and joint structure of neural process models while benefiting from the flexible conditioning capabilities of LLMs.

%NDPs and our LLM-FPs have the advantage that they can learn implicitly how smooth (via the length-scales of the prior samples) the functions they are predicting on are based on the data \citep[demonstrates this, contrasting it with Gaussian Processes]{NDP}; and they can be trained using synthetic data as a prior \citep{transformers-bayesian-inference} for the type of functions that will be learned. The LLM-FP adds the possibility of conditioning on expert knowledge through text data.

\textbf{Future Work.} Several directions could improve LLM-FP itself. 
The NDP component would benefit from scaling and from richer synthetic-data generation procedures \citep{TabPFN2023,tabpfn-nature-v2,DBLP:journals/corr/abs-2502-04573}; linear \citep{DBLP:conf/icml/KatharopoulosV020,DBLP:conf/iclr/ChoromanskiLDSG21} or neighbourhood \citep{DBLP:conf/cvpr/000100LS23,DBLP:journals/corr/abs-2209-15001} attention architectures would reduce the cost of conditioning on large context sets; and text conditioning will naturally improve with stronger LLMs.
The product-of-experts construction itself opens further directions. Our gradient-free convolution method applies to any expert distribution that admits a closed-form Gaussian convolution, not only to LLMs with binned predictions. This makes it applicable to guided optimisation in the style of \citet{DBLP:conf/icml/SongZYM0KCV23,Kong2024-gx}, quantised measurement models, and potentially to guiding diffusion image models with tokenized language-image experts, replacing autoregressive sampling with the marginal-then-couple approach used here.

%\textbf{Future Work.} An area of immediate possible improvement is the further scaling of the NDP models, used and by integrating the use of more recent methods for the generation of complex realistic synthetic data, similar to \citet{TabPFN2023,tabpfn-nature-v2,DBLP:journals/corr/abs-2502-04573}, an area into which we did not invest much effort as it was beyond the scope of our main contributions.
%Computational complexity could be improved through the use of linear attention mechanisms in the flow model architecture.
%Text conditioning of predictions will naturally improve with the integration of more powerful language models.
%Finally, different sampling methods could be used to combine the LLMP and diffusion model experts, for example by adapting the `superposition' method from \citet{DBLP:conf/iclr/SkretaAB0N25} which could potentially reduce cases of over-concentration on regions where NDP and LLMP agree with high density.

%An alternative direction would be the use of our product-of-experts diffusion convolution method in other situations where it can be used to sidestep the usual need for differentiable probabilities.
%For example it could be used in guided optimisation as per \citet{DBLP:conf/icml/SongZYM0KCV23,Kong2024-gx}, or perhaps to guide diffusion image models with tokenized language-image models, replacing the usual autoregression by something akin to the marginal sampling used here by LLMP.

%\todo[inline]{References: should be updated to the inproceedings published versions (try dblp for good bibtex) - checked 5/11.}
\small{\bibliography{references}}

\newpage

\appendix
\onecolumn

\section{Stochastic Processes and Our Models}\label{appendix:KET}

Suppose we have a distribution \(\rho_{x_1, \dots, x_n}(y_1, \dots, y_m)\) indexed and defined for any locations \(x_1, \dots, x_n\) and \(n \ge 1\).
The Kolmogorov Extension Theorem gives sufficient conditions for defining a stochastic process from these as finite dimensional marginals.
The conditions are
\begin{enumerate}
  \item \emph{Exchangeability}: For any permutation \(\pi\) of integers \{1, \dots, n\}, \[ \rho_{x_1, \dots, x_n}(y_1, \dots, y_n) = \rho_{x_{\pi(1)}, \dots, x_{\pi(n)}}(y_1, \dots, y_n).\]
  \item \emph{Consistency}: For any \(1 \le m \le n\), \[ \rho_{x_1, \dots, x_m}(y_1, \dots, y_m) = \int \rho_{x_1, \dots, x_n}(y_1, \dots, y_n) \mathrm{d}y_{m+1} \dots \mathrm{d}y_n.\]
\end{enumerate}

A NDP defined by a diffusion model, the I-LLMP as defined in \citet{LLMP}, and therefore our LLM-FP model all satisfy the \emph{exchangability} property, while the autoregressive A-LLMP does not.
They are all unlikely to satisfy the \emph{consistency} property.
However the NDP could \emph{learn} to be consistent: if it is trained on datasets satisfying consistency, for example Gaussian Process samples, in the limit of perfect learning it exactly models the Bayesian posterior which is consistent.

\section{Flow Matching and Diffusion Additional Details}
\label{appendix:flow-matching}

Flow matching (\citealp{flow-matching-albergo,flow-matching-lipman,flow-straight-fast}; see \citealp{flow-matching-guide} for a comprehensive introduction) generalises the training method for diffusion models \citep{Sohl-DicksteinW15,diffusion-ho,Song}, framing them as a simulation-free training method for continuous normalising flows \citep{neural-ode,neural-ode-ffjord}.
It defines a flow with velocity field \(u_t = \frac{\mathrm{d}\vec{y}_t}{\mathrm{d}t}\) giving rise to a Markov process \((\vec{Y}_t \sim p_t)_{0 \le t \le 1}\) from noise \(p_\text{noise} = p_0\) to data \(p_\text{data} = p_1\).
We would like to learn a parameterised approximation to the velocity field \(u^\theta_t \approx u_t\).
Sampling could then be straightforwardly performed by integration of the ODE defined via this learned velocity field.

The core result of flow matching is that we can replace the ideal flow matching objective \(\mathcal{L}_{FM} = \E_{t, \vec{Y}_t \sim p_t} \| u_t(\vec{Y}_t) - u^\theta_t(\vec{Y}_t)\|^2\) with a version using conditional flows and obtain the same gradients with respect to \(\theta\).
For example, in the `independent coupling' formulation, we can introduce conditional probability \(p_{t|1}(\vec{Y}_t|\vec{Y}_1)\) where integration over \(\vec{Y}_1\) gives the marginals \(p_t\).

From these, we obtain the conditional velocity fields \(u_{t|1}(\cdot|\vec{y}_1)\) which generate the conditional probabilities, and the conditional flow matching objective
\( \mathcal{L}_{CFM} = \E_{t, \vec{Y}_1 \sim p_1, \vec{Y}_t \sim p_{t|1}(\cdot|\vec{Y}_1)} \| u_{t|1}(\vec{Y}_t|\vec{Y}_1) - u^\theta_t(\vec{Y}_t)\|^2.\)

Here we exclusively use the popular class (relating closely to diffusion models) of affine conditional flows \(\vec{Y}_t = \alpha_t \vec{Y}_1 + \sigma_t \vec{Y}_0\), with a Gaussian noise \(p_0\) distribution.
Here the pair \((\alpha_t, \sigma_t)\) is a `scheduler'; these are smooth functions with \(\alpha_0 = \sigma_1 = 0, \alpha_1 = \sigma_0 = 1\), and \(\dot{\alpha}_t, - \dot{\sigma}_t > 0\) on \(t \in (0, 1)\).
This gives conditional probability paths
\[ p_{t|1}(\vec{y}|\vec{y}_1) = \mathcal{N}(\vec{y}|\alpha_t \vec{y}_1, \sigma_t^2 I)\]
and flow matching objective
\[ \E_{t, \vec{Y}_1 \sim p_1, \vec{Y}_t \sim p_{t|1}(\cdot|\vec{Y}_1)} \| \dot{\alpha}_t \vec{Y}_1 + \dot{\sigma}_t \vec{Y}_0 - u^\theta_t(\vec{Y}_t)\|^2. \]
As discussed in \cref{sec:background}, we use the conditional Optimal Transport (OT) path \citep{flow-matching-lipman}: \(\alpha_t = t\) and \(\sigma_t = 1 - t\).

For this special class, due to the Gaussian form of \(p_{t|1}\), the conditional score function can be written as \(\nabla_{\vec{y}} \log p_{t|1}(\vec{y}|\vec{y}_1) = - \frac{1}{\sigma_t^2}(\vec{y} - \alpha_t\vec{y}_1)\).
Through this, the loss function can be further re-parametrised to frame it as learning an approximation to the marginal score functions, \(\nabla_{\vec{y}_t} \log p_t(\vec{y}_t)\), reproducing the de-noising score matching objective \citep{denoising-score-match}.
Other equivalent losses and rescaled variations can also be used, for example predicting the expected final output directly, \(\widehat{\vec{y}}_{1|t}(\vec{y}_t) \coloneq \E[\vec{Y}_1|\vec{Y}_t = \vec{y}_t]\).
These can be linearly converted to the flow formulation via\footnote{
Where we have defined parameters
\begin{align*}
  a_t &\coloneq \frac{\dot{\alpha}_t}{\alpha_t}, &
  b_t &\coloneq \frac{\sigma_t^2 \dot{\alpha}_t - \sigma_t \dot{\sigma}_t \alpha_t}{\alpha_t}, &
  c_t &\coloneq \frac{\dot{\sigma}_t}{\dot{\sigma}_t \alpha_t - \sigma_t \dot{\alpha}_t}, &
  d_t &\coloneq \frac{\sigma_t}{\dot{\sigma}_t \alpha_t - \sigma_t \dot{\alpha}_t}
\end{align*}
derived through manipulations of the affine path (see \emph{e.g.} \citet{flow-matching-guide}).
}
\begin{align}
  u_t(\vec{y}) &= a_t \vec{y} + b_t \nabla_{\vec{y}} \log p_t(\vec{y}) \label{eq:flow-from-score}\\
  \widehat{\vec{y}}_{1|t}(\vec{y}) & = c_t \vec{y} + d_t u_t(\vec{y}) \label{eq:x1-from-flow}.
\end{align}
For OT flow \(u_t(\vec{y}) = \frac{1}{t} \vec{y} + \frac{1-t}{t} \nabla_{\vec{y}} \log p_t(\vec{y})\) and \(\widehat{\vec{y}}_{1|t}(\vec{y}) = \vec{y} + (1-t) u_t(\vec{y})\); this straight path is particularly amenable to discretisation when sampling, motivating the choice.

\Cref{eq:flow-from-score} makes the relationship between flows and learned score functions clear, the resulting ODE exactly matching with the probability flow ODE seen in diffusion models.
To avoid the singularity in \cref{eq:flow-from-score} as \(t \to 0\), we can begin at \(0 < t \ll 1\) where \(p_t\) approximates a Gaussian.

We note that by rearranging \cref{eq:flow-from-score}, \(u_t^\theta\) can be used in the above to define an approximation to the score function at each timestep, \(s_t^\theta\).
This can additionally be used for Langevin steps at any fixed \(t\) (predictor-corrector; \citealp{Song}), giving \emph{stochastic sampling} of the diffusion model; combining a Langevin step with every step of the ODE gives the ``reverse'' stochastic differential equation formulation used by many diffusion models.
Although in principle we could sample directly from the score at the final time step (\emph{i.e.} Langevin sampling on \(p_1\)), this is in practice infeasible, as it is almost impossible to initialise the sampling process in the support of the model (\emph{i.e.} where it has seen training examples), which would lead to poor samples.
This is exactly the problem that diffusion solves,  starting at \(p_0\) or \(p_t\) with \(0 < t \ll 1\) the support of the model is close to Gaussian for which it is trivial to draw initialisations.

\subsection{Relationship of flow matching to diffusion.}\label{sec:fm-vs-diffusion}
Diffusion uses a noising process from data constructed from an stochastic differential equation (SDE) with affine drift coefficients and Brownian motion.
Various motivations lead to objectives like the denoising score matching loss \(\mathcal{L}_{\text{SM}}\).
This happens because we choose an SDE which leads to Gaussian paths \(p_{t|1}\), and thus diffusion can be seen as a special case of conditional flow matching, with different exact SDE formulations leading to different schedulers.
The OT flow is particularly straightforward to learn as it leads to constant velocity flow.
In addition, we note that in the diffusion literature, the parametrisation of time is usually instead defined from data at \(0\) to noise as \(t \to \infty\).

To be specific, diffusion models choose an SDE with affine drift coefficients, and Brownian motion \(\vec{W}_r\),
\[
\mathrm{d}\widetilde{\vec{X}}_r = a_r\widetilde{\vec{X}}_r\,\mathrm{d}r + g_r\mathrm{d}\vec{W}_r,
\quad \widetilde{\vec{X}}_0 \sim p_\text{data}
\quad r \in [0,\infty)
\]
and define \(\vec{X}_{k^{-1}(r)} = \widetilde{\vec{X}}_r\) for some mapping \(k: (0, 1] \to [0, \infty)\); this should be strictly monotonically decreasing with \(k(1) = 0\) and \(\lim_{t \to 0} k(t) = \infty\).
Since \(\vec{X}_0\) is Gaussian, we can define the ``forward'' (using the reversed diffusion time convention) process from data to time \(t\), in closed form as
\( p_{t|1}(\vec{X}_t|\vec{X}_1 = \vec{x}) = \mathcal{N}(\alpha_t \vec{x}, \sigma^2_t I),\)
using coefficients \(\alpha_t\) and \(\sigma_t\) derived from \(a_r\) and \(g_r\) (specifically, as \(\alpha_t = \tilde{\alpha}_r, \sigma_t = \tilde{\sigma}_r\) for \(\tilde{\alpha}_r = \exp \int_0^{r} a_s \mathrm{d}s\) and \(\tilde{\sigma}^2_r = \tilde{\alpha}^2_r \int_0^r \frac{g_s^2}{\tilde{\alpha}^2_s} \mathrm{d}s\)).
The probability flow ODE, corresponding to \cref{eq:flow-from-score}, which runs a ``reverse'' of the above process and has the same marginals \(p_t\) is
\begin{equation}\label{eq:prob-flow-ode}
\mathrm{d}\vec{X}_t = \dot{k}(t) \left[ a_t \vec{X}_t - \frac{g_t^2}{2} \; \nabla_{\vec{X}_t} \log p_t(\vec{X}_t) \right] \mathrm{d}t.
\end{equation}
We can also define a reverse SDE with the same marginals as the original SDE, equivalent to combining probability flow with Langevin dynamics, which is most similar to the original formulation of sampling in diffusion models.

\section{Additional Training Details for NDP}\label{appendix:additional-training}

We parametrise our NDP flow network with a standard diffusion transformer \citep{diffusion-transformer-23}, with the `tokens' or `patches' being concatenated inputs \((\vec{x}^i, \vec{y}^i_t)\) and outputs being the flow corresponding to each \(\vec{y}^i_t\).
As in the above, time is encoded using a sinusoidal embedding scheme and conditioned on through the adaptive Layer Norm-zero (adaLN-zero) scheme. As recommended we initialise the final layer weights to zero.
Since our low-dimensional inputs do not interact well with layer normalisation, these inputs are initially passed through an embedding layer with GeLU \citep{gelu} activation before the first transformer block.

Similarly to \citet{NDP}, we train on a synthetic dataset of Gaussian Process samples of mixed kernel types and length-scales. Data is generated on-the-fly during training so that every minibatch is freshly sampled and the model never sees the same example twice. Each training example draws 256 \((x, y)\) points with \(x \in [-2, 2]\) (sampled stratified-uniform over equal-width bins) and \(y\) supported on \([-3.5, 3.5]\). With equal probability, the GP kernel is drawn from one of two branches:

\textbf{RBF-only}: a squared-exponential kernel with a short length-scale,
\[
k_{\mathrm{rbf}}(x, x') \;=\; \exp\!\left(-\tfrac{1}{2}\,\frac{(x - x')^2}{\ell_{\mathrm{r}}^{2}}\right), \qquad \ell_{\mathrm{r}} \sim \mathcal{U}[0.1,\, 1.0];
\]
\textbf{Locally-periodic}: the product of an exp-sine-squared periodic kernel and a long-scale RBF envelope,
\[
k_{\mathrm{lp}}(x, x') \;=\; \exp\!\left(-\frac{2 \sin^{2}\!\big(\pi (x - x') / p\big)}{\ell_{\mathrm{p}}^{2}}\right) \cdot \exp\!\left(-\tfrac{1}{2}\,\frac{(x - x')^2}{\ell_{\mathrm{R}}^{2}}\right),
\]
with \(p \sim \mathcal{U}[0.75,\, 1.25]\), \(\ell_{\mathrm{p}} \sim \mathcal{U}[0.5,\, 1.0]\), and \(\ell_{\mathrm{R}} \sim \mathcal{U}[1.0,\, 4.0]\).

Both branches use unit signal variance. This 50/50 mixture gives the model both smooth long-trend behaviour and (locally) seasonal structure at test time.
We train the resulting unconditional NDP with an Optimal Transport flow matching objective \citep{flow-matching-albergo,flow-matching-lipman,flow-straight-fast} using AdamW \citep{adamw} for 300,000 steps at batch size 2046 with a constant learning rate of \(3 \times 10^{-4}\).
We do not conduct large-scale scaling, extensive hyper-parameter search, or employ more complex data generation methods which could be used for improved results on real-world data of the NDP alone.
We consider these out-of-scope for demonstrating the utility of our LLM-FP method combining NDPs with LLMPs (for example, \citealp{NDP} also consider other synthetic tasks like learning step functions to demonstrate that non-GP-like function spaces can be modelled, and \citealp{TabPFN2023} consider much more real-data-like structural causal models for data generation).

\subsection{Validation}\label{appendix:validation}

\begin{figure*}[t]
    \centering
    \includegraphics[width=\textwidth]{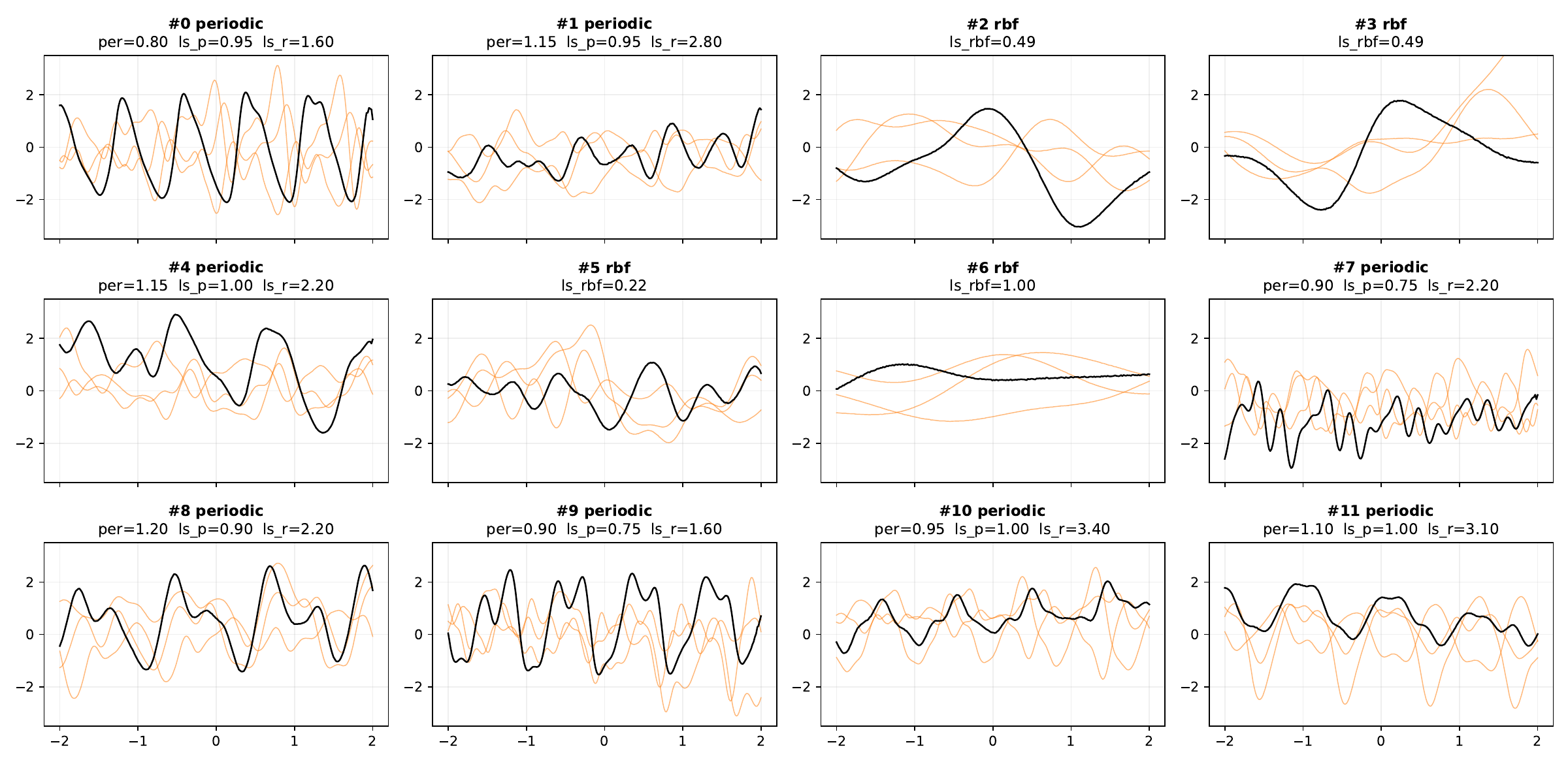}
    \caption{Unconditional NDP samples vs.\ fitted GP samples. Black: unconditional samples drawn from the trained NDP. Orange: samples from a GP whose kernel and hyperparameters were fitted by maximum likelihood to each NDP sample (the inferred kernel branch and hyperparameters are shown above each panel). The two are visually consistent, indicating that individual NDP samples look like plausible GP draws.}
    \label{fig:unconditional_ndp}
\end{figure*}

\begin{figure*}[t]
    \centering
    \includegraphics[width=\textwidth]{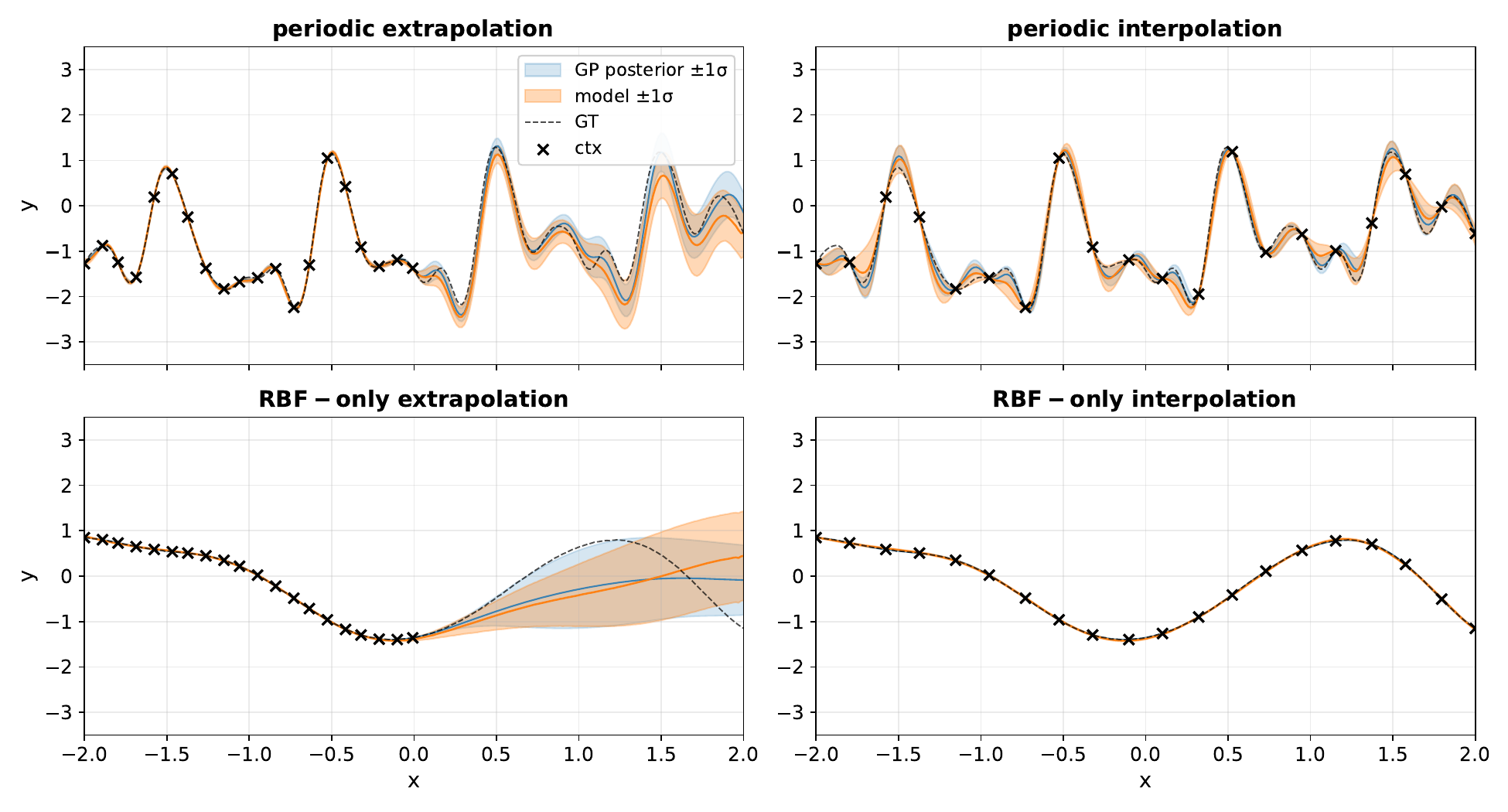}
    \caption{Conditional NDP samples vs.\ GP oracle. For four representative test functions (rows: RBF and locally-periodic), we condition the NDP on a set of context points (black) and show predictive marginals in orange. As an oracle reference, we also draw samples from a GP given the \emph{true} kernel hyperparameters of the test function (blue). The NDP marginalises over its entire training prior, so it cannot match the oracle GP, but its predictive structure is consistent with the ground truth on both interpolation and extrapolation.}
    \label{fig:conditional_ndp}
\end{figure*}

We perform two sanity checks on the trained NDP. \Cref{fig:unconditional_ndp} verifies that unconditional samples are GP-like, and \Cref{fig:conditional_ndp} verifies that conditional samples produce sensible predictive distributions on both interpolation and extrapolation tasks, against a GP oracle baseline.

\section{Computational Costs}
\label{appendix:computational}

NDP model training was conducted on a Nvidia GH200 GPU over about 30 hours.
We estimate an additional 100 hours of preliminary experiments for optimising the NDP configuration.
No LLM training or fine-tuning was conducted.
Inference for all experiments including preliminary ones took about 20 hours of GPU time, overwhelmingly dominated by local LLM inference.

\section{Prompt format}
\label{appendix:prompt}

All LLM calls in our experiments are completions against a chat-templated prompt to Qwen-3.6-27B~\citep{qwen36_27b}. The three LLM-based methods evaluated (\emph{LLM-Independent}, \emph{LLM-Direct}, and the LLM term inside our \emph{LLM-FP} guidance) share the same system prompt and the same way of rendering observed $(x, y)$ data into the user message. They differ only in (a) what the user asks for at the end of the message and (b) whether the assistant first writes a \verb|<think>...</think>| block. Each forecasting task fixes a natural-language description, the numeric range of $y$, the observed context points, and the rendering of $x$ values (integer index, fixed-precision real, or absolute date).

\subsection{System prompt}

The system prompt used in all experiments (with the exception of \Cref{appendix:thinking-ablation}) is identical across problems and methods:

\begin{Verbatim}[breaklines=true,breakanywhere=true,fontsize=\small]
You are a time-series forecasting assistant.

Workflow:
  - You are given a description of the time series and some observed (x, value) pairs.
  - Reason about the function ONCE at the start.
  - You will then be asked to forecast specific x values, one at a time. Reuse your earlier reasoning for each query.

Keep your reasoning to roughly 500 words. You don't need to be exhaustive - sketch the pattern, sanity-check briefly, and stop. If you find yourself re-deriving the same conclusion, enumerating cases, or second-guessing a value you already wrote down, stop and finish.

Internally reason step by step. Briefly cover these points:
  1. The real-world meaning of the values: what units, what is a typical magnitude, what is a plausible upper / lower bound.
  2. The structure of the series: trend (rising / falling), seasonality (daily / monthly / yearly cycle), regime changes (one-time events like a bankruptcy or holiday), or none of the above.
  3. The general shape of the function across the full x range. For seasonal data work out the cycle (for example, if month 1 is February then month 30 is (30 - 1) mod 12 = 5 -> July, which is dry in California).
  4. A plan for how to map any queried x to its forecast value.

Once you have finished reasoning, your visible response is ONLY a single short acknowledgment sentence such as "I have analysed the data." or "Ready to forecast." Do not repeat, summarize, or restate your reasoning in the visible response. Then await the per-x queries. For each query, output ONLY the numeric forecast in the format `x, value`. No commentary, no units, no explanation.
\end{Verbatim}

\subsection{User-message structure}

Every user message has the same four parts:
\begin{enumerate}
  \item A free-form natural-language description of the time series (what the data is, where it comes from, what is expected of it).
  \item A \emph{Constraints} block listing the $y$ range with units and the $x$-domain rule.
  \item The sorted observed $(x, y)$ data lines, one per row, headed with a column label such as \verb|day, Stock Price (GBP)|.
  \item A query line. For per-target scoring the query asks for a single $x_\star$; for full-trajectory sampling it asks for $T$ forecasts at a listed grid of target $x$ values.
\end{enumerate}
$y$ values are rendered with banker's rounding (\texttt{decimal.ROUND\_HALF\_EVEN}) to a task-specific number of decimal places; the same precision determines the spacing of the discrete $y$ grid used by per-target scoring. Context points are always sorted ascending in $x$ before being inserted into the prompt.

\subsection{Date vs.\ numeric \texorpdfstring{$x$}{x}}

How $x$ is rendered depends on the task and falls into three cases:
\begin{itemize}
  \item \textbf{Integer index} (e.g.\ Bankruptcy on day index $0,\dots,50$): the constraint line reads ``Days are integers from $0$ to $50$'' and each data line begins with a bare integer.
  \item \textbf{Fixed-precision real} (e.g.\ Periodic + linear trend on $x \in [-2, 2]$): the constraint line reads ``Xs are real numbers in $[-2.00, 2.00]$'' and $x$ is formatted to a task-specific number of decimal places.
  \item \textbf{Absolute date} (Mauna Loa monthly / weekly): $x$ is emitted as a \verb|YYYY-MM-DD| date, the column header switches to ``date'', the constraint line reads ``Dates are in YYYY-MM-DD (week-start, Sunday-anchored) format covering 2019-01-06 through 2023-11-26.'' (or similar), and the query reads ``Predict the value at 2021-07-11.'' rather than ``Predict the value at week 132.''
\end{itemize}

\subsection{LLM-Independent (per-target scoring)}
\label{appendix:prompt-independent}

For each forecast target $x_\star$ we build the user message described above with the query line ``Predict the value at $x_\star$. Reply in the format \texttt{x, value} with no additional text.'' We then apply Qwen's chat template to the conversation, append the literal string ``$x_\star$, '' to prime the assistant turn up to the comma, and score every candidate $y$ on a regular grid (step matching the $y$-rendering precision) across the prompt $y$ range. The score for each candidate is the sum of the LLM's token log-probabilities for the digits of $y$ followed by a newline. Everything before the candidate $y$ is shared across all candidates, so KV caching reduces each candidate to a short suffix forward pass. The per-grid scores are softmax-normalised over the $y$ grid to give the marginal $\log p_{\mathrm{LLM}}(y \mid x_\star, \mathcal{D}_c)$. \emph{LLM-Independent} samples each $x_\star$ independently from this marginal by inverse-CDF sampling; LLM-FP plugs the same per-target marginal into the guidance term of the NDP flow.

\subsection{LLM-Direct (full-trajectory sampling)}
\label{appendix:prompt-direct}

LLM-Direct queries the model once for the entire forecast vector. The conversation is two user turns: a data-only primer identical to the one used for thinking mode (the assistant emits the \verb|<think>...</think>| block plus a brief acknowledgement); then a single second user turn that explicitly lists the $T$ target $x$ values and demands $T$ lines of output of the form \verb|`x, value`|. Example~A in Section~\ref{appendix:prompt-worked-examples} shows both turns in full for the periodic + linear-trend preset.
We sample $N = 80$ such trajectories with \texttt{do\_sample=True} and constrain the decoder with \texttt{lmformatenforcer} against a regex of the form \verb|<x_1>, -?\d+(?:\.\d+)?\n<x_2>, -?\d+(?:\.\d+)?\n...\n|, where the \verb|<x_i>| are the literal target $x$ values from the prompt, so that every completion is guaranteed to parse into a length-$T$ vector with the prescribed $x$ values in the correct order.

\subsection{Thinking mode}
\label{appendix:prompt-thinking}

For thinking-enabled runs we generate the \verb|<think>...</think>| block exactly once per evaluation (per task and per random seed) by passing \verb|enable_thinking=True| to the chat template, then re-use it across every per-target query (LLM-Independent, LLM-FP) or in front of the second user turn (LLM-Direct). Qwen-3.6 supports preserving a previously-emitted \verb|<think>| block across subsequent turns via a \verb|preserve_thinking=True| flag, which we use to share the thinking across queries. This sharing is what makes per-target scoring tractable: the model reasons about the whole function once (up to 4096 tokens of internal reasoning) and the per-target queries pay only the cost of scoring the $y$ grid against a cached prefix.

\subsection{Worked examples}
\label{appendix:prompt-worked-examples}

\paragraph{Example A --- Periodic + linear trend (real-valued $x$, LLM-Direct).} Examples A and B illustrate two different methods on two different tasks: A shows LLM-Direct with full-trajectory sampling, B shows LLM-Independent / LLM-FP with the multi-turn thinking-mode path. The data-only first user turn is structurally the same in both cases. The preset for Example A uses 20 context points evenly spaced on $x \in [-2, 0]$ at full \texttt{float64} resolution (true spacing $2/19 \approx 0.10526$, which renders as alternating $0.10$ / $0.11$ steps at a 2-decimal-place rendering). The LLM is asked to forecast on the full domain $x \in [-2, 2]$ at $T = 51$ target locations, obtained by subsampling 51 of the NDP's 256 dense grid points via \texttt{np.linspace(0, 255, 51).round()}; the targets are therefore approximately but not exactly uniformly spaced (see Appendix~\ref{appendix:alignment} for the alignment constraint motivating this). In LLM-Direct mode the first user turn is data-only --- it primes the model to think about the function but does not yet ask for any forecast value:

\begin{Verbatim}[breaklines=true,breakanywhere=true,fontsize=\small]
I have a time series forecasting task for you. Here's some context about the task:

A periodic function with a linear trend

Numerical constraints:
- Values are real numbers in [-3.5, 3.5].
- Xs are real numbers in [-2.00, 2.00].

Observed data (x, value):
-2.00, -1.69
-1.89, -0.54
-1.79, -0.04
-1.68, 0.52
-1.58, 0.09
-1.47, 0.20
-1.37, -0.95
-1.26, -1.24
-1.16, -0.70
-1.05, -0.76
-0.95, 0.50
-0.84, 0.57
-0.74, 1.01
-0.63, 0.76
-0.53, 0.24
-0.42, 0.22
-0.32, -0.37
-0.21, -0.10
-0.11, 0.18
0.00, 1.16

Think carefully about the function: its general shape across the full range, any seasonality, trends, regime changes, or events. After you have thought you do not need to respond, just acknowledge you are ready.

Each forecast you give will be in the format `x, value`.
\end{Verbatim}

\noindent
The assistant then emits its single \verb|<think>...</think>| block followed by a one-sentence acknowledgement. A second user turn then requests all $T = 51$ forecast values in one go:

\begin{Verbatim}[breaklines=true,breakanywhere=true,fontsize=\small]
Now produce the full forecast.

Forecast at the following xs (in this exact order):
-2.00, -1.92, -1.84, -1.76, -1.69, -1.59, -1.51, -1.44, -1.36, -1.28, -1.20, -1.12, -1.04, -0.96, -0.89, -0.81, -0.71, -0.64, -0.56, -0.48, -0.40, -0.32, -0.24, -0.16, -0.09, -0.01, 0.09, 0.16, 0.24, 0.32, 0.40, 0.48, 0.56, 0.64, 0.71, 0.79, 0.89, 0.96, 1.04, 1.12, 1.20, 1.28, 1.36, 1.44, 1.51, 1.59, 1.69, 1.76, 1.84, 1.92, 2.00

Reply with EXACTLY 51 lines and nothing else. Each line must be `x, value` --- the x from the list above, then a comma and a space, then the predicted value as a real number. Output the 51 lines in the same order as the list. Do not include headers, commentary, units, blank lines, or any text before or after the 51 lines.
\end{Verbatim}

\paragraph{Example B --- Mauna Loa weekly CO\textsubscript{2}, 20 random context weeks (date $x$, LLM-Independent / LLM-FP, thinking mode).} Same machinery, with $x$ rendered as Sunday-anchored weekly dates and a custom date-range constraint line. To prevent memorisation, these dates are chosen at random and the corresponding CO\textsubscript{2} values are interpolated from the monthly data (\emph{cf.}~the subsection on memorisation in \citealp{context-is-key}). In thinking mode the per-target query is a \emph{multi-turn} conversation: a data-only primer (the same form as Example~A, just with the Mauna Loa task description and observations), one assistant turn that contains the \verb|<think>...</think>| block plus a brief ``Ready.'' acknowledgement, and then \emph{one short user turn per target $x_\star$}.

\begin{Verbatim}[breaklines=true,breakanywhere=true,fontsize=\small]
I have a time series forecasting task for you. Here's some context about the task:

Weekly mean atmospheric CO2 concentration in ppm at Mauna Loa Observatory across roughly 5 years (256 weekly observations from 2019-01-06 through 2023-11-26). The provided observations are 20 sparsely-and-randomly sampled weeks from the first half (through 2021-06-13); forecast the remainder.

Numerical constraints:
- Values represent CO2 (ppm), real numbers in [405.0, 430.0].
- Dates are in YYYY-MM-DD (week-start, Sunday-anchored) format covering 2019-01-06 through 2023-11-26.

Observed data (date, CO2 (ppm)):
2019-01-13, 411.0
2019-02-03, 411.6
2019-03-03, 412.1
2019-05-26, 414.6
2019-08-04, 410.8
2019-09-01, 409.4
2020-02-02, 414.1
2020-03-01, 414.6
2020-04-26, 416.8
2020-05-10, 417.2
2020-05-17, 417.2
2020-06-07, 416.7
2020-06-28, 415.6
2020-07-12, 414.7
2020-09-27, 411.5
2020-10-11, 411.5
2020-11-01, 412.5
2021-02-07, 416.6
2021-04-18, 419.0
2021-06-13, 418.9

Think carefully about the function: its general shape across the full range, any seasonality, trends, regime changes, or events. After you have thought you do not need to respond, just acknowledge you are ready.

Each forecast you give will be in the format `date, value`.
\end{Verbatim}

\noindent
The assistant then emits a \verb|<think>...</think>| block (which we share across all 64 per-target queries) followed by a one-sentence acknowledgement (e.g.\ ``Ready to forecast.''). For \emph{each} forecast date $x_\star$ we then append a single user turn of the form
\begin{Verbatim}[breaklines=true,breakanywhere=true,fontsize=\small]
Forecast at date=2021-07-11.
\end{Verbatim}
\noindent
re-render the whole conversation through the chat template with \verb|preserve_thinking=True|, \verb|enable_thinking=False|, and prime the assistant turn with ``\verb|2021-07-11, |'' so the next token is the first digit of $y$. LLM-Independent / LLM-FP construct the per-target marginal $\log p_{\mathrm{LLM}}(y \mid x_\star, \mathcal{D}_c)$ by scoring $y$ on a $0.1$-ppm grid spanning $[405, 430]$ against this prefix. KV caching means the system prompt, primer, and shared assistant-with-\verb|<think>| turn are computed only once, and each per-target query reuses that cache --- only the short ``\verb|Forecast at date=...|'' suffix is recomputed per $x_\star$.

The per-target queries are issued at 64 dates subsampled from the 256-week NDP grid at evenly spaced indices, spanning the full $x$ range and not just the forecast region: 32 sit inside the context window and 32 sit in the forecast region (\verb|2021-07-11|, \verb|2021-08-08|, \dots, \verb|2023-11-26|). The LLM-FP guidance therefore conditions the NDP at 64 anchor points. The corresponding LLM-Direct prompt is the same primer followed by the \emph{single} batched user turn shown in Section~\ref{appendix:prompt-direct}, requesting all 64 dates at once.

\section{LLM--NDP target alignment}
\label{appendix:alignment}

The LLM-FP score correction in \cref{eq:moe-score} couples the LLM marginal $\log \widetilde{q}_{r_t}(\widehat{\vec{y}}_{1|t} \mid c)$ to the NDP's expected denoised output $\widehat{\vec{y}}_{1|t}$ at the same target locations. Concretely, if the LLM is queried at $x$ value $x_\star$ but the NDP is sampling at a different set of $x$ values, there is no $\widehat{\vec{y}}_{1|t}$ to evaluate the LLM marginal against, and the guidance term is undefined.

In practice, our NDP transformer takes $x$ values as part of its input tokens and can in principle predict at any chosen set of locations, so this is not a fundamental constraint. We choose to align the LLM target locations with a subset of the NDP's sampling grid for two reasons. First, it lets the LLM marginal at $x_\star$ enter the score function for the NDP's prediction at exactly that $x_\star$, without interpolation between LLM anchor points. Second, querying the LLM at every NDP grid point would be expensive (the per-target scoring cost is dominated by the number of LLM queries); aligning a sparser subset of LLM targets to a denser NDP grid lets us trade off LLM cost against NDP resolution. The 51 LLM targets in Example~A (see \Cref{appendix:prompt-worked-examples}) are obtained by subsampling the NDP's 256-point grid at uniform indices via \texttt{np.linspace(0, 255, 51).round()}; the 64 LLM dates in Example~B are obtained the same way.

Predictions and metrics are computed at the full NDP grid (256 points in both cases). The LLM-FP guidance therefore conditions the NDP at the LLM-anchored subset, and the NDP's joint structure determines values at the unanchored positions through its score function.

\section{Thinking vs no thinking}
\label{appendix:thinking-ablation}

All results in this paper use thinking-enabled prompts, where the LLM emits a \verb|<think>...</think>| reasoning block before forecasting (Section~\ref{appendix:prompt-thinking}). For completeness we examine the effect of disabling thinking, replacing the per-target query path described in Section~\ref{appendix:prompt-independent} with the data-only single-turn variant and dropping the \verb|<think>| block.

\begin{figure*}[t]
    \centering
    \includegraphics[width=\textwidth]{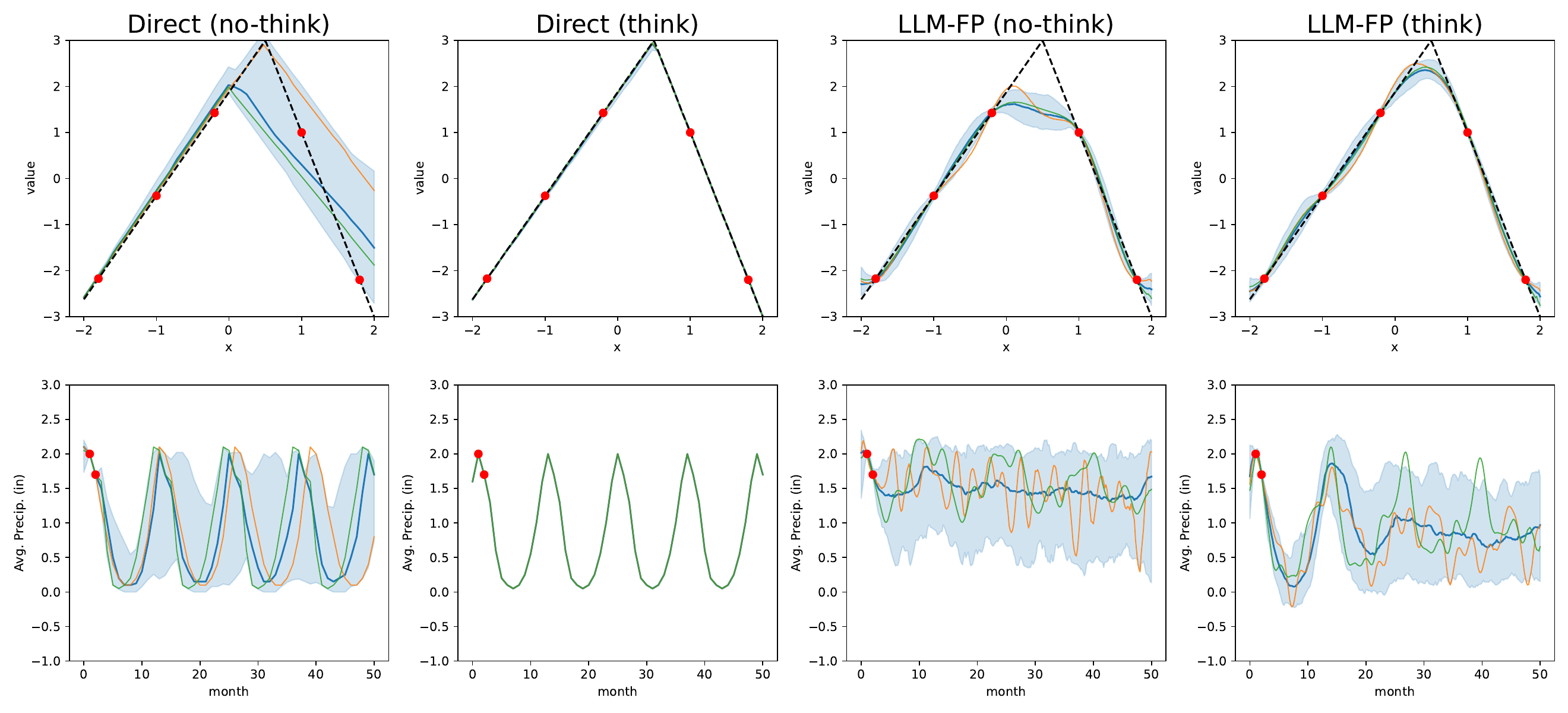}
    \caption{Effect of enabling LLM reasoning ("thinking") on the Change point (top) and San Diego (bottom) tasks. Reasoning generally improves consistency and predictive accuracy across the LLM-based methods, but can also reduce predictive uncertainty: on San Diego, LLM-Direct with reasoning collapses to near-identical samples.}
    \label{fig:think_vs_nothink}
\end{figure*}

On Change point, both LLM-Direct and LLM-FP fail to capture the discontinuity without thinking: LLM-Direct picks linear continuations that miss the true function, and LLM-FP smooths out the peak at the intersection of the two segments. With thinking, both recover the change-point structure.

San Diego shows a different trade-off. Without thinking, LLM-Direct produces samples that each individually repeat a fixed pattern but vary across samples, giving better predictive diversity than the thinking-enabled version, where all samples collapse onto a single periodic template. LLM-FP without thinking retains almost no structure, with samples reverting to the NDP prior; with thinking it concentrates on more plausible trajectories.

Thinking helps when the model can lock onto structure (Change point) and can hurt when it leaves the autoregressive rollout with insufficient residual variability (LLM-Direct on San Diego). LLM-FP is less exposed to the latter: per-target marginals retain more variability than autoregressive rollouts even after a confident reasoning pass, for the reasons discussed in \cref{sec:discussion}.

\section{Additional Experiments and Details}\label{sec:further-experimental}

\subsection{Metrics Definitions}
\label{appendix:metric-definitions}

We report three metrics for every method on every task: mean absolute error (MAE), continuous ranked probability score (CRPS) in its empirical-ensemble form, and negative log predictive density (NLPD) under a Gaussian fit to the ensemble. Each method produces an ensemble of $N$ trajectories $\{y^{(i)}\}_{i=1}^{N}$, where each $y^{(i)} \in \mathbb{R}^{T}$ is a sample over the evaluation grid $x_1, \dots, x_T$ used for that task (the size and extent of which depends on whether the task uses forecast-region or full-grid scoring; see below), and the ground truth is a deterministic function $y^\star(x)$. We use $N = 800$ for NDP, LLM-FP, and LLM-Independent and $N = 80$ for LLM-Direct (as it is substantially more expensive per sample). All three metrics are reported in the original $y$ units of each task (no normalisation) and are oriented so that lower values indicate better forecasts. For most tasks, the average is taken over the \emph{forecast region only}, the subset of grid positions with $x > \max(\mathcal{D}_c)$, i.e.\ those positions strictly beyond the rightmost context observation. The exception is the change-point task, for which we report metrics averaged over the full evaluation grid as method's behaviour on either side of the change point is informative.

\paragraph{MAE.} The MAE of the per-position ensemble mean against the truth,
\[
  \mathrm{MAE} \;=\; \frac{1}{T} \sum_{t=1}^{T} \big| \bar{y}_t - y^\star(x_t) \big|, \qquad \bar{y}_t \;=\; \frac{1}{N} \sum_{i=1}^{N} y^{(i)}_t.
\]
This summarises the point-forecast accuracy of the ensemble's centre and reduces to ordinary MAE for a deterministic predictor.

\paragraph{CRPS.} The empirical-ensemble form of CRPS \citep{gneiting2007strictly},
\[
  \mathrm{CRPS} \;=\; \frac{1}{T} \sum_{t=1}^{T} \left[ \frac{1}{N} \sum_{i=1}^{N} \big| y^{(i)}_t - y^\star(x_t) \big| \;-\; \frac{1}{2 N^{2}} \sum_{i=1}^{N} \sum_{j=1}^{N} \big| y^{(i)}_t - y^{(j)}_t \big| \right].
\]
This is a strictly proper score that rewards both calibration and sharpness, and collapses to MAE when the ensemble is a single repeated point.

\paragraph{NLPD.} For each evaluation position $t$ we fit a Gaussian $\mathcal{N}(\mu_t, \sigma_t^{2})$ to the ensemble and report the average negative log density of the ground-truth value under that fit:
\[
  \mathrm{NLPD} \;=\; \frac{1}{T} \sum_{t=1}^{T} \left[ \tfrac{1}{2} \log\!\big(2 \pi \sigma_t^{2}\big) \;+\; \frac{\big(y^\star(x_t) - \mu_t\big)^{2}}{2\, \sigma_t^{2}} \right],
\]
where
\[
  \mu_t \;=\; \frac{1}{N} \sum_{i=1}^{N} y^{(i)}_t, \qquad \sigma_t \;=\; \sqrt{\frac{1}{N - 1} \sum_{i=1}^{N} \big(y^{(i)}_t - \mu_t\big)^{2}}.
\]
We use the Bessel-corrected (\texttt{ddof=1}) sample standard deviation. The logarithm is natural, so NLPD is reported in nats. To prevent divergence on positions where the ensemble has collapsed to a near-deterministic point, we clip $\sigma_t$ from below at $\sigma_{\min} = 10^{-3}$ in the original $y$ units before evaluating the formula. This is a numerical guard, not a model assumption: it caps the spurious $\log \sigma$ savings of a degenerate ensemble while still penalising confidently-wrong predictions through the $(y^\star - \mu)^{2} / 2 \sigma_{\min}^{2}$ term.

\paragraph{Uncertainty estimation.} Each cell in the results table is reported as $m_{\pm h}$, where $m$ is the metric computed as defined above and $h = (q_{97.5} - q_{2.5})/2$ is half the width of the 95\% bootstrap confidence interval. The bootstrap resamples the $T$ evaluation positions of the cell with replacement, repeats this $B = 1000$ times, and takes the $2.5\%$ and $97.5\%$ quantiles of the resulting metric distribution. This captures uncertainty arising from \emph{which} test points the metric was averaged over rather than Monte Carlo noise from the predictive ensemble, which a separate sample-bootstrap shows to be typically $3$--$15\times$ tighter at $N = 800$ and is therefore not reported.

\subsection{Additional qualitative comparisons}
\label{appendix:additional-tasks}

\begin{figure*}[t]
    \centering
    \includegraphics[width=\textwidth]{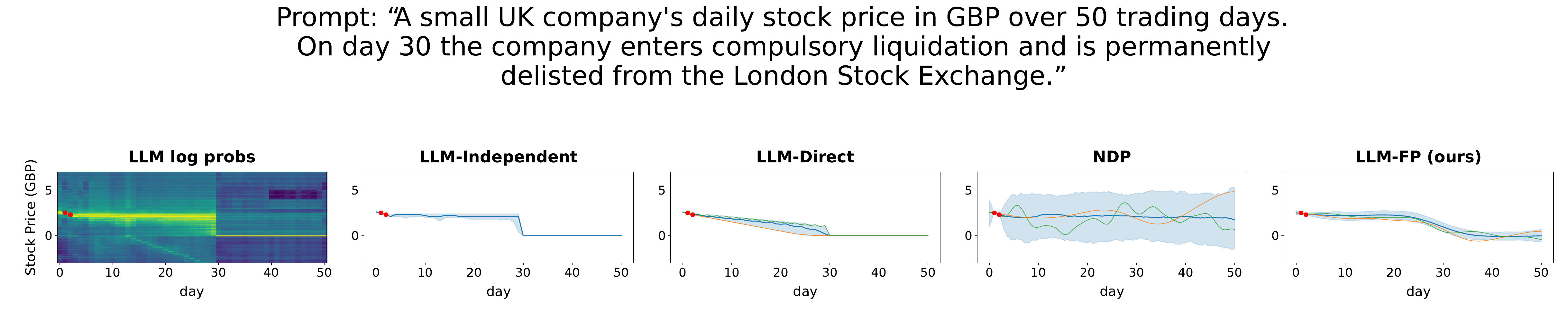}
    \caption{Synthetic bankruptcy scenario. The text-conditioned methods follow the day-30 constraint specified in the prompt; the NDP extrapolates from its training prior.}
    \label{fig:bankruptcy}
\end{figure*}

\begin{figure*}[t]
    \centering
    \includegraphics[width=\textwidth]{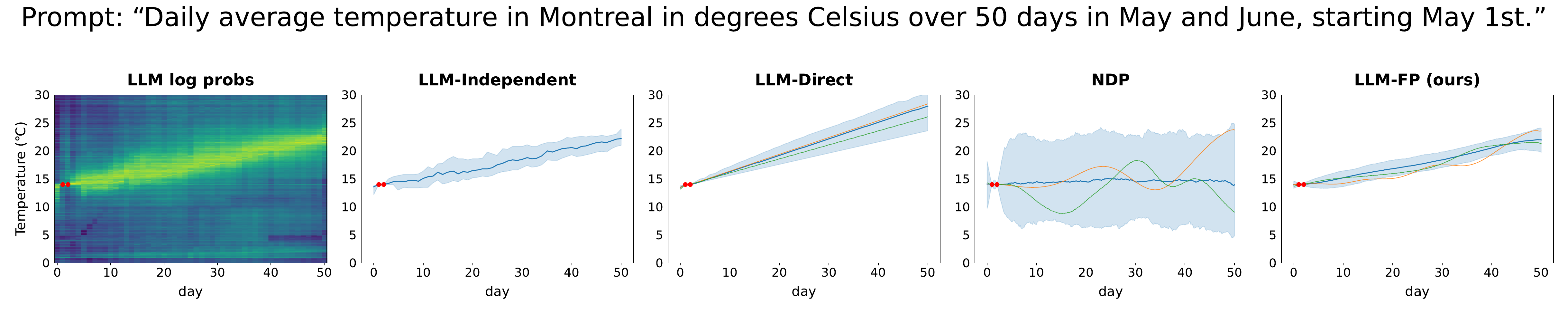}
    \caption{Daily temperature extrapolation in Montreal. All text-conditioned methods produce reasonable predictions; LLM-Direct samples tend to follow near-linear trajectories rather than the seasonal structure suggested by the LLM marginals.}
    \label{fig:montreal}
\end{figure*}

\Cref{fig:bankruptcy,fig:montreal} show two additional tasks not included in the main paper: a synthetic bankruptcy scenario where the prompt specifies that the value goes to zero at day 30, and Montreal daily temperature extrapolation with a textual description of the location and variable.

Bankruptcy is a sharp test of textual conditioning: the day-30 constraint is unrecoverable from the numerical context alone. All text-conditioned methods (LLM-Direct, LLM-Independent, LLM-FP) respect the constraint, while the unconditional NDP extrapolates from its training prior and ignores it.

Montreal is a softer setting where all text-conditioned methods produce broadly reasonable predictions. LLM-Direct samples tend to follow near-linear trajectories not matching those suggested by the LLM marginals; LLM-FP combines the LLM marginals with the NDP's local structure to produce smoother seasonal trajectories.

\end{document}